\documentclass[a4paper,fleqn]{cas-dc}

\usepackage{comment}
\usepackage{amsmath,amssymb,amsfonts}
\usepackage{algorithmic}
\usepackage{graphicx}
\usepackage{textcomp}
\usepackage{booktabs}
\usepackage{xcolor}
\usepackage{url}
\usepackage[table]{xcolor}
\usepackage{float}
\usepackage{ulem}
\usepackage{hyperref}

\usepackage[numbers]{natbib}

\def\tsc#1{\csdef{#1}{\textsc{\lowercase{#1}}\xspace}}
\tsc{WGM}
\tsc{QE}

\definecolor{lightblue}{RGB}{230,242,255}

\begin{document}

\makeatletter
\def\ps@first{%
  \let\@oddhead\@empty
  \let\@evenhead\@empty
  \let\@oddfoot\@empty
  \let\@evenfoot\@empty
}
\def\ps@cas{%
  \let\@oddhead\@empty
  \let\@evenhead\@empty
  \let\@oddfoot\@empty
  \let\@evenfoot\@empty
}
\makeatother

\pagestyle{plain}

\shorttitle{Parameter-Efficient Vision–Language Adaptation with Continuous Metadata Conditioning for Animal Re-Identification}    


\title [mode = title]{Parameter-Efficient Vision–Language Adaptation with Continuous Metadata Conditioning for Animal Re-Identification}  

\tnotemark[1] 

\tnotetext[1]{}

\author[1]{Anil Osman Tur}[
  orcid=0000-0001-7772-5235
]
\ead{anilosman.tur@univr.it}

\author[2,3]{Tonje Knutsen Sørdalen}[
  orcid=0000-0001-5836-9327
]
\ead{tonje.sordalen@hi.no}

\author[2]{Kim Tallaksen Halvorsen}[
  orcid=0000-0001-6857-2492
]
\ead{kim.halvorsen@hi.no}

\author[1]{Cigdem Beyan}[
  orcid=0000-0002-9583-0087
]
\cormark[1]
\ead{cigdem.beyan@univr.it}

\affiliation[1]{organization={Department of Computer Science, University of Verona},
            city={Verona},
            postcode={37134},
            country={Italy}}

\affiliation[2]{organization={Institute of Marine Research, Nye Flødevigveien 20},
            postcode={4817 His}, 
            country={Norway}}

\affiliation[3]{organization={University of Agder, Centre for Coastal Research},
            city={Grimstad},
            postcode={4604},
            country={Norway}}

\cortext[1]{Corresponding author. Email: cigdem.beyan@univr.it}

\fntext[1]{}


\begin{abstract}
Long-term animal re-identification (ReID) must remain robust to gradual morphological evolution and seasonal appearance shifts. Although recent vision–language models provide strong pretrained visual representations, adapting them to longitudinal ecological settings remains challenging, particularly under identity and temporal distribution shifts.  
We present a parameter-efficient CLIP adaptation framework for animal ReID and introduce a continuous metadata-conditioning mechanism that incorporates numerical attributes directly into the prompt representation during training. While low-rank visual adaptation, prompt-based supervision, and cross-modal alignment provide the adaptation framework, the proposed metadata-conditioning strategy constitutes the primary methodological contribution. By preserving the continuous structure of numerical metadata rather than discretizing it into textual categories, the proposed approach enables smooth modulation of the embedding space during training while maintaining a purely visual inference pipeline.
Experiments on a seven-year longitudinal fish dataset and multiple wildlife benchmarks demonstrate improved performance under closed-set, open-set, and time-aware evaluation protocols. The results demonstrate that continuous metadata conditioning improves robustness to longitudinal appearance variation and temporal distribution shifts, while parameter-efficient adaptation enables a purely visual inference pipeline without requiring metadata at test time. Code and evaluation splits can be found at: \url{https://github.com/AnilOsmanTur/MetaPrompt-ReID}.
\end{abstract}




\maketitle


\begin{center}
\noindent
\small
\textbf{Author's Note.}
This is the author's accepted manuscript of the paper accepted for publication in
\textit{Expert Systems with Applications}.
The final authenticated version will be available from the publisher.
\end{center}

\section{Introduction}
\label{sec:intro}

Animal re-identification (ReID) refers to the recognition of previously observed individuals across time and has long been an important component of wildlife population monitoring, capture–recapture studies, and behavioral ecology~\cite{mcclintock2014mark}. In computer vision, the term more specifically denotes identifying individual animals from images or video sequences without the use of invasive tagging. Automated visual ReID enables large-scale longitudinal tracking under natural conditions~\cite{huang2025uniformity,compte2025housed,liu2024fishtrack}; however, this requires models to learn identity-discriminative representations capable of matching individuals across encounters based solely on visual appearance as it changes over time~\cite{nepovinnykh2020siamese,ravoor2020survey,li2020atrw,Beyan2026}.

Unlike person or vehicle ReID, animal ReID frequently involves long-term re-observation, where the same individual may be encountered months or even years apart~\cite{adam2024seaturtleid,nepovinnykh2022sealid,bai2024amurtiger}. This setting introduces challenges beyond viewpoint and illumination variation. Animals exhibit deformable body structures and substantial pose variability, while inter-individual differences are often subtle and fine-grained. Moreover, appearance evolves over time due to growth, seasonal coloration shifts, and life-stage transitions, increasing intra-individual variation and complicating identity matching across extended temporal spans~\cite{cermak2024wildfusion,ravoor2020survey,jiao2023toward,Beyan2026}.

In this work, we focus on fish ReID, a setting that is ecologically and economically relevant for fisheries management and population monitoring~\cite{canovi2024trajectory,olsen2023contrastive}. We conduct experiments on the Melops dataset~\cite{sordalen2025melopsreid,Sordalen2026wild}, which contains images of wild corkwing wrasse (\textit{Symphodus melops}) collected over seven years. The dataset comprises approximately 24K images of 9K PIT-tagged individuals and includes around 3K resighting events spanning multiple years.

Melops~\cite{sordalen2025melopsreid,Sordalen2026wild} reflects a fully in-the-wild data collection setting, with images acquired and annotated within a biological monitoring framework. As such, it captures realistic longitudinal variability rather than curated short-term observations. The dataset exhibits several characteristics not fully represented in many existing animal ReID benchmarks. First, it contains substantial longitudinal variation, as individuals undergo natural growth and seasonal appearance changes over multi-year time spans. Second, the identity distribution is highly imbalanced, with many individuals observed only once and a smaller subset repeatedly re-sighted. Third, it provides rich capture-level metadata, including body length, morph type, spawning status, spatial location, and precise capture date. These properties make Melops a challenging and ecologically realistic benchmark for studying long-term identity stability and metadata-aware representation learning.

Recent advances in vision-language models (VLMs), particularly CLIP~\cite{radford2021learning}, provide a strong pretrained foundation for ReID by learning transferable visual representations from large-scale image-text data. Such large-scale pretraining has the potential to improve robustness to subtle inter-individual differences and gradual appearance changes, which are common in longitudinal animal monitoring settings. Recent CLIP-based ReID approaches adapt these pretrained encoders by introducing identity-specific learnable text tokens and leveraging cross-modal alignment losses. In particular, CLIP-ReID~\cite{li2023clip} adopts a two-stage training strategy in which ID-specific text tokens are first optimized while keeping both encoders frozen, and the image encoder is subsequently fine-tuned under constraints imposed by the learned text representations. Similarly, IndivAID~\cite{wu2024individual} adopts a two-stage framework in which a text description generator first produces individual- and image-specific textual descriptions, and an attention module subsequently merges these descriptions to guide the fine-tuning of the image encoder.

While these approaches demonstrate the benefit of exploiting CLIP's cross-modal structure for ReID, they rely on staged optimization procedures and additional components to mediate interaction between visual and textual features. In longitudinal animal ReID, where identity must remain stable while appearance evolves over extended time spans, it is desirable to adapt pretrained representations in a stable and parameter-efficient manner without substantially expanding the architecture. 
Moreover, in many ecological monitoring programs, structured numerical attributes accompany visual observations, including capture time, body size, or reproductive state. Such attributes are biologically meaningful and may correlate with systematic appearance variation over time. Existing methods, e.g.,~\cite{li2025metawild,adam2024seaturtleid}, typically incorporate metadata by converting it into discrete textual descriptions or by integrating it through additional attention-based fusion mechanisms. However, incorporating continuous numerical metadata directly within a learnable prompt representation, without introducing additional expert branches, cross-attention modules, or inference-time metadata dependencies, remains largely unexplored in animal ReID.

In this work, we propose a parameter-efficient adaptation framework for animal ReID built upon a frozen CLIP backbone. 
Our method jointly optimizes low-rank visual adaptation (LoRA~\cite{lora2022}) modules in the vision encoder together with learnable prompt context tokens, while retaining fixed template tokens. Unlike staged optimization strategies, both visual adaptation and prompt parameters are trained end-to-end under a unified objective. Cross-modal contrastive alignment between projected visual features and identity-conditioned text embeddings further structures the shared embedding space during training. 
At inference time, all text-related components and metadata inputs are discarded, and ReID is performed via nearest-neighbor retrieval using visual descriptors.
This design enables deployment without inference-time metadata or auxiliary modules while preserving the general visual knowledge of the pretrained model and adapting it through a limited number of additional trainable parameters.

To address longitudinal appearance variation, we further introduce continuous metadata conditioning. Instead of discretizing numerical attributes into symbolic text tokens, we preserve their numerical structure by injecting them directly into the learnable prompt representation. This allows metadata to induce smooth geometric modulation of the embedding space, which is better aligned with metric learning objectives than step-wise symbolic discretization. We investigate sinusoidal encodings and FiLM-based modulation mechanisms to condition the prompt during training, allowing metadata to influence representation learning without introducing additional fusion branches or attention modules.

We evaluate our approach under four protocols: closed-set, open-set, time-aware closed-set, and time-aware open-set, designed to reflect realistic deployment scenarios. Across these settings, our method consistently improves performance over CLIP-based baselines. We observe notable gains in performance under both closed-set and open-set conditions, and metadata conditioning further enhances performance in temporally constrained evaluation. Additional experiments on multiple benchmarks, including various animals other than fish, confirm consistent generalization across species and acquisition conditions.

While our framework builds on established components such as LoRA~\cite{lora2022} and prompt learning, the primary methodological contribution of this work lies in a fundamentally different paradigm for incorporating metadata in ReID. Rather than treating metadata as discrete tokens or auxiliary modalities, we formulate it as a continuous training-time conditioning signal that shapes the geometry of the embedding space. This enables smooth adaptation to longitudinal appearance variation and, critically, decouples representation learning from inference requirements by eliminating any dependence on metadata at test time. In summary, our contributions are as follows.

\begin{itemize}

\item \textbf{Continuous metadata-conditioned prompt learning.}
We introduce a metadata-conditioning mechanism that embeds numerical metadata directly into the prompt representation, avoiding discretization and eliminating the need for auxiliary fusion modules. This design enables smooth modulation of the embedding space and better aligns with the gradual temporal and morphological changes inherent in longitudinal animal ReID.

\item \textbf{Training-time-only metadata conditioning without inference dependency.}
We propose a formulation in which metadata is used exclusively during training to shape the representation space, while all metadata and text-related components are removed at inference time. This preserves a purely visual deployment pipeline, in contrast to prior multimodal approaches that rely on metadata during inference.

\item \textbf{Unified end-to-end parameter-efficient adaptation framework.}
We develop a single-stage training strategy that jointly optimizes low-rank visual adaptation, learnable prompt context tokens, and symmetric cross-modal contrastive alignment under a frozen CLIP backbone. Unlike prior staged or multi-component pipelines, our approach enables stable and efficient adaptation without full backbone fine-tuning.

\item \textbf{Longitudinal evaluation protocol and temporal analysis.}
We introduce time-aware evaluation protocols that explicitly model temporal distribution shifts and provide an empirical analysis of identification performance as a function of temporal distance, highlighting the impact of longitudinal appearance variation.

\item \textbf{Comprehensive empirical validation across datasets.}
We demonstrate consistent improvements over CLIP-based baselines across multiple evaluation settings and wildlife benchmarks, while using substantially fewer trainable parameters and maintaining strong generalization performance.

\end{itemize}

The remainder of this paper is organized as follows. 
Sec.~\ref{sec:related} reviews related work in animal ReID, longitudinal wildlife datasets, VLMs in animal ReID, and parameter-efficient adaptation strategies. 
Sec.~\ref{sec:method} presents the proposed framework, including low-rank visual adaptation, prompt learning, and continuous metadata conditioning. 
Sec.~\ref{sec:experiments} describes the experimental setup, evaluation protocols, implementation details, and empirical results. 
Sec.~\ref{sec:discussions} discusses the findings and their implications for longitudinal ecological ReID, and Sec.~\ref{sec:conclusion} concludes the paper.


\section{Related Work}
\label{sec:related}

\subsection{Animal ReID}
\label{subsec:animalREID}

Although automated ReID has been extensively studied in the context of person and vehicle analysis~\cite{li2018richly,quan2019auto,zhu2020voc,khan2019survey,zhang2025diffusion}, animal ReID has more recently emerged as an important and rapidly growing research area. It plays a critical role in wildlife monitoring, ecological studies, and livestock management by enabling non-invasive identification of individuals across images and videos~\cite{ravoor2020survey,Beyan2026,huang2025uniformity,liu2024fishtrack,compte2025housed}.

A fundamental distinction between animal ReID and person or vehicle ReID lies in the temporal and observational assumptions underlying the task.
Person and vehicle ReID are typically formulated in multi-camera settings, where the same individual is observed across different viewpoints within relatively short time spans, making viewpoint variation and camera-specific bias the dominant challenges~\cite{Beyan2026}.
In contrast, many animal ReID scenarios operate under fixed or semi-fixed acquisition setups, e.g.,~\cite{borlinghaus2023bumblebees,murali2019classification}, but may involve re-observation of the same individual over weeks or years apart~\cite{adam2024seaturtleid, nepovinnykh2022sealid}.
While only a subset of existing studies explicitly address such long-term ReID, animal ReID must additionally contend with deformable body structures, substantial pose variation, and fine-grained inter-individual differences~\cite{he2023posture,bai2024amurtiger,cermak2024wildlifedatasets}.
Consequently, animal ReID places greater emphasis on learning robust identity representations from intrinsic appearance cues, rather than relying primarily on short-term cross-camera or viewpoint transitions.

With the widespread adoption of deep learning techniques, most animal ReID approaches rely on convolutional neural networks (CNNs) trained with metric learning objectives, such as triplet or contrastive loss~\cite{li2020atrw,cheng2020defat,moskvyak2021mantaray}.
To improve robustness under pose deformation and background clutter, many works incorporate pose-aware or part-based representations, attention mechanisms, or geometric constraints, either through explicit keypoints and landmarks~\cite{li2020atrw,moskvyak2020landmark} or detection-driven masking strategies~\cite{cheng2020defat,he2023posture}.
More recently, transformer-based architectures and hybrid CNN–Transformer models have been explored to capture long-range dependencies and multi-granularity features, demonstrating improved performance~\cite{zheng2022transformer,li2024redeformtr,bai2024amurtiger,lamping2025chickens}.

Beyond deep metric learning, a distinct line of research adopts content-based image retrieval (CBIR) principles for animal ReID.
These methods avoid end-to-end metric learning and instead rely on local pattern descriptors, geometric consistency, and similarity matching to identify individuals~\cite{nepovinnykh2024norppa,nepovinnykh2024species,nepovinnykh2025multiimage}.
Such CBIR-based pipelines are particularly effective for patterned species and support open-set identification without retraining; however, they often depend on accurate segmentation, handcrafted or semi-handcrafted descriptors, and species-specific design choices, which can limit scalability and adaptability across datasets.

Recent studies have also explored incorporating auxiliary cues to enhance discriminability in animal ReID.
These include pose information, often obtained through keypoint or landmark estimation and used to guide feature extraction or alignment~\cite{moskvyak2020landmark,li2020atrw,ROSENBERG2026645};
behavioral metadata, such as standing or lying states, which condition identity matching on activity context~\cite{perneel2025cattle,he2023posture};
orientation information derived from body geometry or spine direction to normalize viewpoint effects~\cite{perneel2025cattle};
temporal context from video sequences to aggregate identity evidence across frames or tracklets~\cite{williams2025track,yu2025holstein};
and segmentation masks that isolate animal regions or distinctive patterns before feature extraction~\cite{nepovinnykh2024norppa,cheng2020defat}.
These auxiliary cues are typically fused with visual features to mitigate appearance ambiguity and improve robustness.
While such multimodal strategies can yield performance gains, they often rely on additional annotations (e.g., pose keypoints, behavior labels, segmentation masks) or task-specific modules, such as pose estimators, behavior classifiers, detection-driven attention masks, or geometry-based alignment components.
As a result, these approaches are frequently tailored to particular species, datasets, or acquisition protocols, which can limit scalability and generalization across animal ReID scenarios.

Overall, existing animal ReID methods, whether based on deep metric learning or CBIR, primarily rely on visual supervision and species-specific modeling, with limited exploitation of large-scale pretrained priors. In contrast, vision–language models offer a complementary direction by leveraging semantic and cross-modal knowledge learned from massive image–text corpora.
However, the application of such models to animal ReID remains relatively unexplored.

\subsection{Recent Animal Re-ID Datasets}
Animal Re-ID datasets have expanded in scale and annotation richness, shifting from small benchmarks annotated only with identity labels toward ecologically grounded datasets that provide complementary supervision, such as localization, pose, and temporal metadata~\cite{cermak2024wildlifedatasets,jiao2023toward}. Despite this progress, most datasets remain temporally shallow and do not capture long-term appearance evolution in a biologically meaningful sense.

Early wildlife benchmarks such as ATRW~\cite{li2020atrw} established Re-ID feasibility under unconstrained conditions through rich spatial and pose annotations, but rely on short-term collections and relatively few individuals due to ecological constraints. A similar pattern appears in outdoor pasture and farm datasets, including YakReID-103~\cite{zhang2021yakreid} and the Holstein dairy cow dataset~\cite{li2021individual}, which demonstrate robustness to environmental variability but lack longitudinal identity tracking.

Only a small number of datasets extend beyond short-term observations. SealID~\cite{nepovinnykh2022sealid} aggregates multi-year data and supports fine-grained pattern matching through segmentation and pelage annotations, although identity observations remain sparse and seasonally clustered. SeaTurtleID2022~\cite{adam2024seaturtleid} represents a notable exception, spanning 13 years and introducing time-aware closed-set (identity overlap with temporal separation) and open-set (temporally emerging unseen individuals) evaluation protocols, while also demonstrating that random splits substantially overestimate performance. Nevertheless, for long-lived species, such as sea turtles, even decade-scale data capture only a limited fraction of lifespan-related appearance drift.

Other benchmarks primarily incorporate short-term temporal cues via videos or tracklets rather than true long-term monitoring. For example, MultiCamCows2024~\cite{yu2025holstein} and PolarBearVidID~\cite{zuerl2023polarbearvidid} provide multi-camera observations and motion information over days or short sequences, supporting cross-view and motion-aware Re-ID but remaining temporally shallow. As a result, reported temporal robustness in animal Re-ID often reflects short-term appearance consistency rather than genuine long-term identity stability.

In contrast to the benchmarks discussed above, the fish dataset used in this paper provides a relatively large-scale longitudinal setting specifically designed to study long-term identity stability under natural growth and seasonal variation. Unlike short-term or cross-view datasets, it captures continuous morphological changes associated with indeterminate growth, as well as seasonal coloration shifts and life-stage transitions. Moreover, the dataset exhibits a highly imbalanced identity distribution, where the majority of individuals are observed only once while a subset is repeatedly recaptured over extended intervals.

\subsection{Fish ReID Methods and Datasets}
Visual ReID of individual fish is increasingly studied due to its relevance for fisheries management, behavioral ecology, and population monitoring. Fish present distinct challenges for long-term ReID, as many species exhibit indeterminate growth with continuous changes in body size and shape, while coloration and pigmentation may vary with age, season, social status, or environmental conditions~\cite{price2008pigments,john2021coloration,uglem2000phenotypic}. These biological factors undermine appearance constancy assumptions and complicate temporally robust identification.

Early fish ReID approaches relied on classical computer vision methods and focused on species with distinctive surface patterns. Pattern-based matching of spots, stripes, or scars has been widely applied to whale sharks and manta rays, combining local keypoint descriptors with manual or semi-automated verification~\cite{holmberg2009population,town2013mantamatcher}. Such methods are effective at moderate temporal scales, typically months to a few years, provided that visual patterns remain stable and image quality is sufficient.

Recent work increasingly adopts deep learning, typically using CNNs trained with metric learning objectives. For instance,~\cite{gomezvargas2023undulateskate} demonstrates few-shot Siamese learning. Lonati et al.~\cite{lonati2024epaulette} further show that growth-induced appearance changes significantly affect deep ReID performance across life stages in epaulette sharks.

Despite these advances, fish ReID remains limited by the scarcity of datasets with repeated observations of known individuals over long time spans. Most datasets feature few individuals, sparse re-sightings, or short temporal coverage, restricting evaluation of long-term robustness. Addressing this issue,~\cite{olsen2023contrastive} shows that fish facial patterns can remain identifiable over multi-year intervals, although performance degrades as temporal gaps increase. Overall, the combination of fine-grained inter-individual variation, continuous growth, and temporal appearance change makes fish ReID a particularly challenging problem.

\subsection{Vision–Language Models for Animal ReID}
\label{subsec:VLM}

Recent work has explored VLMs as a way to provide structured cross-modal supervision for Re-ID, even when only identity labels are available, and no explicit semantic descriptions are given.
CLIP-ReID~\cite{li2023clip} is the first work to adapt CLIP to the Re-ID setting, demonstrating that identity discrimination can be achieved by learning identity-specific text tokens in the absence of concrete textual labels. While effective, the method relies on a two-stage optimization procedure in which prompt tokens and image encoders are optimized sequentially, preventing fully joint end-to-end training.

Several follow-up approaches extend CLIP-based Re-ID to animal domains characterized by stronger appearance variability. Wu et al.~\cite{wu2024individual} propose an identity-driven framework, called IndivAID, that generates image- and individual-specific textual descriptions, which are then used to refine visual embeddings. Although this design improves robustness to pose and viewpoint variation, it also introduces a multi-stage pipeline with frozen components and additional attention modules, increasing training complexity and limiting end-to-end optimization. Jiao et al.~\cite{jiao2023toward} further incorporate large language models (LLMs) to guide semantic prompt construction for cross-species and open-world animal Re-ID. While such LLM-assisted designs improve generalization to unseen categories, they depend on substantially larger models than CLIP and external prompt-generation mechanisms, raising concerns regarding efficiency, reproducibility, and practical deployment.

Beyond language-only supervision, some work has explored multimodal augmentation of VLM-based Re-ID. MetaWild~\cite{li2025metawild}, for example, integrates environmental metadata through adapter-style modules to improve robustness under ecological variability. Although effective, this approach treats metadata as an auxiliary modality that must be explicitly modeled through additional architectural components and assumes the availability of such metadata during representation learning.

Overall, existing VLM- and MLLM-based Re-ID methods demonstrate the promise of language and auxiliary information for identity discrimination, but they commonly rely on staged optimization, auxiliary modules, external language models, or increased model scale. In addition, auxiliary information is typically incorporated either by converting numerical metadata into discrete textual descriptions or by introducing dedicated fusion mechanisms, which may impose artificial discretization boundaries or increase architectural complexity.

In contrast, our approach embeds continuous numerical attributes directly into the prompt representation, preserving their numerical structure and enabling smooth geometric modulation of the embedding space. Furthermore, metadata is used exclusively during training and is not required at inference time, resulting in a purely visual deployment pipeline without additional dependencies. This form of continuous metadata conditioning within the prompt space, combined with training-time-only usage without inference dependency, has not been explored in the context of ReID.

\subsection{Parameter-Efficient Fine-Tuning and Prompt Learning}
\label{subsec:PEFT}
PEFT has emerged as an effective strategy for adapting large pretrained models while avoiding the cost and overfitting risks of full fine-tuning~\cite{houlsby2019parameter,zaken2022bitfit}. Representative approaches include adapter modules, partial layer fine-tuning, low-rank adaptations, and prompt-based methods, all of which update only a small subset of parameters while keeping the backbone largely frozen. Among these, Low-Rank Adaptation (LoRA)~\cite{lora2022} has gained particular attention due to its simplicity and effectiveness, injecting trainable low-rank updates into existing layers without introducing additional inference-time overhead.

Prompt learning provides a complementary PEFT mechanism by adapting the input representation rather than the model weights. Context Optimization (CoOp)~\cite{zhou2022coop} demonstrates that learning continuous prompt tokens can effectively adapt frozen VLMs to downstream tasks, while follow-up works explore variations such as class-specific prompts, deeper prompt insertion, or hybrid prompt-adapter designs~\cite{zhou2022coop,jia2022visual}. Despite their success, different prompt formulations and PEFT strategies often exhibit dataset- and task-dependent behavior, and their relative effectiveness remains an empirical question~\cite{lester2021power}.

In this work, we adopt LoRA-based visual adaptation together with learnable prompt context tokens, enabling joint end-to-end optimization under a frozen CLIP backbone. We further provide a systematic ablation study comparing different PEFT configurations and prompt-learning strategies.

\begin{figure*}[tb!]
    \centering
    \includegraphics[width=\textwidth]{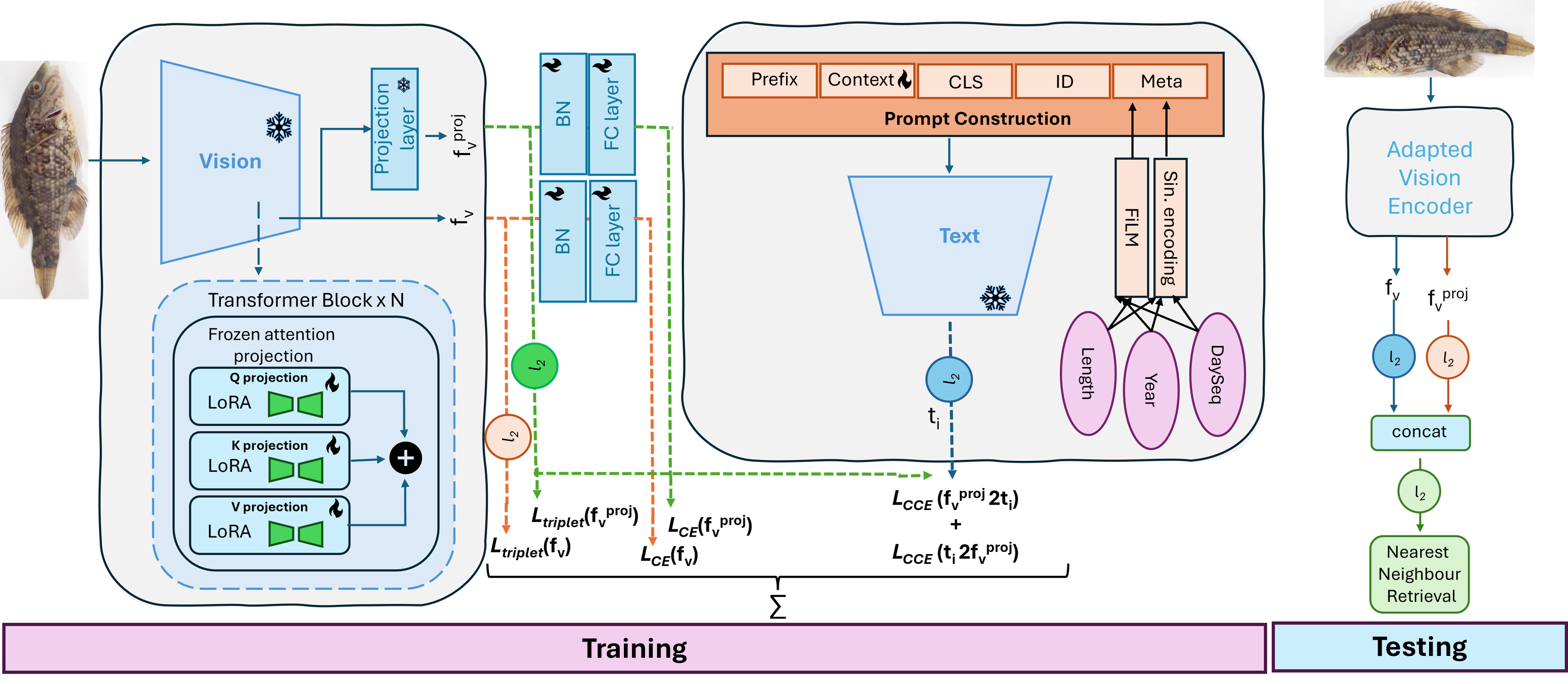}
    \caption{Overview of the proposed framework.
\textbf{Training (left):} A frozen CLIP ViT vision encoder is adapted using LoRA modules inserted into the query, key, and value linear projection layers of each multi-head self-attention block. The visual embedding $\mathbf{f}_v$ and its CLIP-projected counterpart $\mathbf{f}_v^{\mathrm{proj}}$ are supervised with dual-space batch-hard triplet losses and auxiliary identity classification heads. In parallel, identity-conditioned prompts composed of fixed prefix tokens, learnable context tokens, class tokens, identity tokens, and optional metadata tokens are processed by a frozen CLIP text encoder to produce text embeddings $\mathbf{t}$. Continuous metadata is incorporated via sinusoidal encoding or FiLM-based modulation at the embedding level. Cross-modal alignment between $\ell_2$-normalized projected visual features $\mathbf{f}_v^{\mathrm{proj}}$ and text embeddings $\mathbf{t}$ is enforced using symmetric contrastive cross-entropy losses $\mathcal{L}_{\mathrm{CCE}}(\mathbf{f}_v^{\mathrm{proj}}, \mathbf{t})$ and $\mathcal{L}_{\mathrm{CCE}}(\mathbf{t}, \mathbf{f}_v^{\mathrm{proj}})$.
\textbf{Inference (right):} Only the LoRA-adapted vision encoder is retained. The $\ell_2$-normalized embeddings $\mathbf{f}_v$ and $\mathbf{f}_v^{\mathrm{proj}}$ are concatenated and re-normalized to form the final descriptor used for nearest-neighbor retrieval. All text-related components, metadata conditioning, and auxiliary identity heads are removed at inference time.}
    \label{fig_method_sum}
\end{figure*}

\section{Proposed Method}
\label{sec:method}

Given a training dataset of image-identity pairs 
$\{(\mathbf{x}_i, y_i)\}_{i=1}^N$, 
where $\mathbf{x}_i \in \mathbb{R}^{H \times W \times 3}$ denotes an input image and 
$y_i \in \{1, \ldots, C\}$ represents the corresponding identity label, 
the goal of Re-ID is to learn an embedding function 
$f : \mathcal{X} \rightarrow \mathbb{R}^{d}$ 
that maps images of the same identity close together while separating different identities in a shared feature space. 
To this end, we propose a parameter-efficient adaptation framework built upon a frozen vision-language backbone, 
in which identity-aware visual representations are learned via the joint optimization of low-rank visual adaptation modules, 
learnable prompt-based supervision, and metadata-conditioned prompt representations. Fig.~\ref{fig_method_sum} provides an overview of the proposed framework.

The combination of low-rank visual adaptation and prompt-based conditioning is motivated by their complementary roles: LoRA enables controlled adaptation of pretrained visual features, while prompt-based supervision structures the embedding space through cross-modal alignment. Continuous metadata conditioning further extends this framework by introducing smooth, biologically meaningful variation aligned with temporal drift.

\subsection{Architecture Overview}
\label{subsec:overview}
Our framework builds upon the CLIP model~\cite{radford2021learning} and adopts a Vision Transformer (ViT) as the image encoder. 
The vision encoder produces visual features of dimension $d_v$, which are further projected into the shared embedding space of dimension $d_t$, consistent with the CLIP text embedding space.

Given an input image $\mathbf{x}$, the vision encoder first partitions the image into non-overlapping patches and maps them to patch embeddings. 
Let $N_p$ denote the number of image patches. 
A learnable class token is prepended to the patch sequence, and positional embeddings are added, yielding the initial token representation 
$\mathbf{z}_0 = [\mathbf{z}_{\mathrm{cls}};\, \mathbf{E}_{\mathrm{patch}}(\mathbf{x})] + \mathbf{E}_{\mathrm{pos}}$, 
where $\mathbf{z}_{\mathrm{cls}} \in \mathbb{R}^{d_v}$ denotes the class token, 
$\mathbf{E}_{\mathrm{patch}}(\mathbf{x}) \in \mathbb{R}^{N_p \times d_v}$ is the patch embedding projection, 
and $\mathbf{E}_{\mathrm{pos}} \in \mathbb{R}^{(N_p+1) \times d_v}$ represents positional embeddings. 
The resulting token sequence $\mathbf{z}_0 \in \mathbb{R}^{(N_p+1) \times d_v}$ is processed by a stack of $L$ transformer blocks.

As commonly adopted in ViT architectures, the final visual representation is obtained from the class token of the last transformer layer, 
$\mathbf{f}_v = \mathbf{z}_L^{\mathrm{cls}}$. 
This representation is subsequently projected into the CLIP embedding space via a linear projection 
$\mathbf{f}_v^{\mathrm{proj}} = \mathbf{f}_v \mathbf{W}_{\mathrm{proj}}$, 
where $\mathbf{W}_{\mathrm{proj}} \in \mathbb{R}^{d_v \times d_t}$ denotes the pretrained projection matrix used to map visual features into the shared CLIP embedding space. 
The projected embeddings are $\ell_2$-normalized before metric learning and retrieval.

All pretrained backbone parameters, including both the vision encoder and the text encoder, remain frozen during training. 
Model adaptation is achieved through additional lightweight modules introduced in subsequent sections.

\subsection{Low-Rank Adaptation (LoRA)}
\label{subsec:lora}

To enable parameter-efficient adaptation while preserving the pretrained knowledge of CLIP, we introduce LoRA~\cite{lora2022} modules into the ViT-based vision encoder while keeping all original backbone parameters frozen. 
Specifically, LoRA is applied to the query, key, and value linear projection layers within each multi-head self-attention (MHA) block, allowing the model to adapt to identity-level discrimination without full fine-tuning.

Let $\mathbf{W}_0 \in \mathbb{R}^{d_{\mathrm{out}}\times d_{\mathrm{in}}}$ denote a pretrained linear weight (for attention projections, $d_{\mathrm{in}}$ and $d_{\mathrm{out}}$ correspond to the layer input and output dimensions; in practice $d_{\mathrm{in}}=d_{\mathrm{out}}=d_v$ for the per-layer projections unless otherwise noted). 
LoRA augments the forward computation as $\mathbf{h} = \mathbf{W}_0 \mathbf{x} + \frac{\alpha}{r}\mathbf{B}\mathbf{A}\mathbf{x}$, where $\mathbf{A}\in\mathbb{R}^{r\times d_{\mathrm{in}}}$ and $\mathbf{B}\in\mathbb{R}^{d_{\mathrm{out}}\times r}$ are trainable low-rank matrices with $r \ll \min(d_{\mathrm{out}}, d_{\mathrm{in}})$, and $\alpha$ is a scalar scaling factor controlling the magnitude of the adaptation. 
The pretrained weight $\mathbf{W}_0$ (and any associated bias) remains fixed throughout training; only $\mathbf{A}$ and $\mathbf{B}$ are optimized.
This formulation is equivalent to applying an additive low-rank update to the pretrained weight, i.e., $\mathbf{W} = \mathbf{W}_0 + \frac{\alpha}{r}\mathbf{B}\mathbf{A}$, while keeping $\mathbf{W}_0$ fixed.
In practice, LoRA is applied to the full projection matrices prior to head partitioning within MHA.

In our experiments, $\mathbf{A}$ and $\mathbf{B}$ are initialized such that the LoRA branch is effectively inactive at initialization (e.g., initializing $\mathbf{A}$ to zero or using small-scale initialization for $\mathbf{B}$), thereby preserving pretrained behavior at the start of training. This design enables efficient adaptation to the ReID task while retaining the general visual knowledge encoded in the frozen CLIP backbone.

\subsection{Learnable Prompt-Based Text Representation}
\label{subsec_prompt}

To adapt the frozen CLIP text encoder to the ReID task, we adopt the Context Optimization (CoOp) framework~\cite{zhou2022coop} and introduce a small set of learnable context tokens within a fixed prompt template. Following CoOp, the text encoder remains fully frozen, and only the context tokens are optimized to adapt the pretrained CLIP model to the target domain.

Each prompt is composed of four types of tokens: 
(i) fixed prefix tokens corresponding to the textual phrase ``this is a photo of'', 
(ii) shared learnable context tokens, 
(iii) a fixed class token that denotes the semantic category (e.g., \texttt{[fish]}), and 
(iv) a fixed identity token. 

The identity token is defined per identity and shared across all images of the same individual; different identities are assigned distinct identity tokens, enabling identity-specific conditioning of the prompt representation. 
Each identity token is implemented as a randomly initialized embedding vector in the CLIP text embedding space and remains fixed throughout training. These tokens are not part of the pretrained CLIP vocabulary and are used solely to construct identity-conditioned prompts during training. Although identity tokens may appear superficially similar to proxy-based or prototype-based representations used in metric learning \cite{kim2020proxy,biehl2009metric}, they serve a fundamentally different role. Proxy-based methods introduce learnable class representatives in the visual embedding space that act as classification anchors. In contrast, identity tokens in our framework are fixed, non-learnable embeddings defined in the text (prompt) space. They do not function as class prototypes, but instead provide identity-specific conditioning that facilitates cross-modal alignment during training.

All prompt components, including prefix, context, class, identity, and metadata tokens introduced below, are represented directly as embedding vectors in the CLIP text embedding space and concatenated prior to being processed by the frozen text encoder. The resulting embedding sequence is passed to the frozen text transformer, with positional encodings applied in the standard CLIP manner.

To reduce sensitivity to prompt structure, we construct multiple prompt variants by permuting the position of the learnable context tokens relative to the fixed class and identity tokens, while preserving the relative order of the class and identity tokens.
Specifically, we consider three prompt layouts:
\begin{itemize}
    \item \textbf{End}: \([\mathbf{e}_{\mathrm{prefix}},\, \mathbf{e}_{\mathrm{cls}},\, \mathbf{C},\, \mathbf{e}_{\mathrm{id}}^{\,i}]\),
    \item \textbf{Middle}: \([\mathbf{e}_{\mathrm{prefix}},\, \mathbf{C}_1,\, \mathbf{e}_{\mathrm{cls}},\, \mathbf{C}_2,\, \mathbf{e}_{\mathrm{id}}^{\,i}]\),
    \item \textbf{Front}: \([\mathbf{e}_{\mathrm{prefix}},\, \mathbf{C},\, \mathbf{e}_{\mathrm{cls}},\, \mathbf{e}_{\mathrm{id}}^{\,i}]\),
\end{itemize}
where $\mathbf{C} \in \mathbb{R}^{M \times d_t}$ denotes the full set of $M$ learnable context tokens shared across identities, and $\mathbf{C}_1$ and $\mathbf{C}_2$ correspond to the first and second halves of $\mathbf{C}$, respectively. 
In all cases, the class token precedes the identity token, but they are not required to be adjacent.

Each prompt variant is processed by the frozen CLIP text encoder to produce a layout-specific embedding. 
The final identity-conditioned text representation is obtained by averaging the $\ell_2$-normalized embeddings across all prompt variants:
\begin{equation}
\mathbf{t}_i = \frac{1}{|\mathcal{K}|} \sum_{k \in \mathcal{K}}
\frac{\mathrm{TextEncoder}(\mathbf{p}_i^{(k)})}
{\left\lVert \mathrm{TextEncoder}(\mathbf{p}_i^{(k)}) \right\rVert_2},
\label{eq:text_embedding_avg}
\end{equation}
where $\mathcal{K}$ indexes the set of prompt layouts. 
Averaging normalized embeddings across prompt variants encourages invariance to prompt structure while preserving semantic consistency in the shared CLIP embedding space.

During training, only the shared context tokens $\mathbf{C}$ are optimized jointly with the LoRA parameters in the vision encoder, while the text encoder weights, prefix tokens, class tokens, and identity tokens remain fixed. 
Although the text branch is discarded at inference time, the identity-conditioned prompts provide structured supervision during training, encouraging the vision encoder to learn visual embeddings that are geometrically aligned with identity-aware textual representations in the shared CLIP embedding space.

\subsection{Metadata-Conditioned Prompt Adaptation}
\label{subsec:metadataInject}

In addition to identity labels, longitudinal ecological monitoring programs systematically record structured numerical metadata, such as capture date, body length, spatial location, and reproductive status. These attributes are not auxiliary in a statistical sense, but reflect biologically meaningful processes, including growth, seasonal coloration, and life-stage transitions that directly influence visual appearance. 

Unlike prior approaches that discretize such information into categorical bins or fixed tokens (e.g.,~\cite{li2025metawild}), we preserve its continuous structure and incorporate it directly into the prompt representation. Discretization may discard fine-grained numerical relationships and introduce artificial boundaries that limit generalization, particularly in open-world or longitudinal settings where newly observed values may fall outside predefined bins. We therefore encode real-valued attributes as continuous embeddings compatible with the CLIP text embedding space, without quantization. By operating directly in the embedding space, metadata can facilitate smoother adaptation of the learned representation geometry. This design more naturally reflects the gradual nature of longitudinal appearance variation and supports metric learning under temporal drift.

Metadata is used only during training, where it acts as a conditioning signal that shapes the geometry of the learned embedding space. By exposing the model to continuous temporal and morphological variation, the visual encoder learns to internalize these patterns within its representations. As a result, the model becomes more robust to such variations at inference time, even in the absence of metadata, which is particularly beneficial under temporal distribution shifts and open-set conditions.

Let $\mathbf{a} \in \mathbb{R}^{n_a}$ denote a vector of numerical metadata associated with an input image. 
The metadata is image-specific and may vary across different samples of the same identity. 
In all cases, a metadata embedding $\mathbf{e}_a \in \mathbb{R}^{d_t}$ is appended to the prompt sequence after the identity token, yielding prompts of the form 
$[\mathbf{e}_{\mathrm{prefix}},\, \mathbf{C}^{(\cdot)},\, \mathbf{e}_{\mathrm{cls}},\, \mathbf{e}_{\mathrm{id}}^{\,i},\, \mathbf{e}_a]$, 
where $\mathbf{C}^{(\cdot)}$ denotes one of the prompt layouts defined in Sec.~\ref{subsec_prompt}. 
The position of the metadata token remains fixed and is not affected by the layout permutations described previously.

We introduce two alternative strategies for constructing the metadata embedding $\mathbf{e}_a$: 
(i) sinusoidal encoding followed by linear projection, and 
(ii) Feature-wise Linear Modulation (FiLM)-based conditioning of learnable attribute tokens. 
Both approaches produce embeddings in $\mathbb{R}^{d_t}$ that are appended to the prompt sequence without altering the architecture of the frozen CLIP text encoder.

\subsubsection{Sinusoidal Metadata Encoding}
Inspired by positional encodings in transformer models, we encode each scalar component of the numerical metadata using sinusoidal functions at multiple frequencies. 
For each metadata dimension $a_k$, we compute 
$\mathrm{PE}(a_k)_{2j} = \sin(\omega_j a_k)$ and 
$\mathrm{PE}(a_k)_{2j+1} = \cos(\omega_j a_k)$, 
where $\{\omega_j\}$ are fixed frequencies spaced exponentially following the standard Transformer formulation. 
The encodings for all metadata dimensions are concatenated to form $\mathrm{PE}(\mathbf{a})$.

The resulting sinusoidal representation is projected into the CLIP text embedding space via a learnable linear transformation $\mathbf{W}_a \in \mathbb{R}^{d_t \times d_{\mathrm{PE}}}$ and normalized:
$\mathbf{e}_a = \mathrm{BN}(\mathbf{W}_a\,\mathrm{PE}(\mathbf{a}))$,
where $\mathrm{BN}(\cdot)$ denotes batch normalization. 

The metadata embedding $\mathbf{e}_a \in \mathbb{R}^{d_t}$ is computed from the input metadata through the parameterized transformation described above and is not treated as a learnable prompt token.

\subsubsection{FiLM-Based Metadata Conditioning}
As an alternative, we adopt Feature-wise Linear Modulation (FiLM)~\cite{perez2018film} to condition prompt representations on numerical metadata.
In this formulation, the attribute vector $\mathbf{a} \in \mathbb{R}^{n_a}$ is mapped to feature-wise scaling and shifting parameters via a learnable transformation 
$\mathbf{W}_{\mathrm{film}} : \mathbb{R}^{n_a} \rightarrow \mathbb{R}^{2d_t}$:
$[\boldsymbol{\beta}, \boldsymbol{\gamma}] = \mathrm{BN}(\mathbf{W}_{\mathrm{film}}(\mathbf{a}))$,
where $\boldsymbol{\beta}, \boldsymbol{\gamma} \in \mathbb{R}^{d_t}$ denote additive and multiplicative modulation vectors, respectively.
Batch normalization is applied to stabilize the scale of the generated modulation parameters.

These parameters are applied feature-wise to a set of learnable attribute basis tokens $\mathbf{T}_a \in \mathbb{R}^{M_a \times d_t}$, shared across samples, as
$\mathbf{e}_a = \boldsymbol{\gamma} \odot \mathbf{T}_a + \boldsymbol{\beta}$,
where modulation is broadcast across the $M_a$ tokens.
This allows metadata-dependent modulation of attribute-specific basis tokens while operating entirely at the embedding level prior to the frozen CLIP text encoder.

For both sinusoidal and FiLM-based variants, metadata embeddings are computed per image during training and appended to the prompt sequence as standard token embeddings. 
These embeddings receive positional encodings and participate in self-attention without modification to the CLIP text encoder architecture.

At inference time, metadata and all text-related components are removed, and ReID relies solely on visual embeddings. 
Thus, metadata serves as a training-time conditioning signal that shapes the embedding geometry without introducing inference-time dependencies.

\subsection{Auxiliary Identity Supervision}
\label{subsec:auxiliary}

Although identity tokens provide structured supervision through the text branch, they do not constitute a classification mechanism in the visual embedding space. To encourage explicit identity separation in the visual representation, we introduce auxiliary identity supervision directly on the features learned by the vision encoder.

Auxiliary supervision is particularly beneficial in our setting, where most of the CLIP backbone is frozen, and the model is adapted in a parameter-efficient manner using low-rank updates and prompt learning. While the primary objective is metric learning via triplet loss, such relative constraints can be noisy and sensitive to batch composition, especially in early training stages. Cross-entropy supervision provides an absolute class-level signal that stabilizes optimization and promotes compact intra-identity clusters.

Specifically, given the visual representation $\mathbf{f}_v \in \mathbb{R}^{d_v}$ extracted from the final transformer layer and its projection $\mathbf{f}_v^{\mathrm{proj}} \in \mathbb{R}^{d_t}$ in the shared CLIP embedding space, we apply batch normalization followed by linear classifiers:
$\mathbf{s}_v = \mathbf{W}_{\mathrm{aux}}^{(v)} \mathrm{BN}(\mathbf{f}_v),\ 
\mathbf{s}_p = \mathbf{W}_{\mathrm{aux}}^{(p)} \mathrm{BN}(\mathbf{f}_v^{\mathrm{proj}})$,
where $\mathbf{W}_{\mathrm{aux}}^{(v)} \in \mathbb{R}^{C \times d_v}$ and $\mathbf{W}_{\mathrm{aux}}^{(p)} \in \mathbb{R}^{C \times d_t}$ are trainable weights. The resulting logits are supervised using cross-entropy loss with identity labels. Supervising both feature spaces ensures that identity discrimination is enforced both before and after the frozen CLIP projection.

The auxiliary identity heads are used only during training and removed at inference time.

\subsection{Loss Function}
\label{subsec:loss}

The model is trained using a composite objective that combines (a) metric learning, (b) cross-modal alignment, and (c) auxiliary identity supervision.

Given projected visual embeddings $\mathbf{f}_{v,i}^{\mathrm{proj}}$ and identity-conditioned text embeddings $\mathbf{t}_i$, we compute the similarity matrix
\begin{equation}
S_{ij} =
\frac{\mathbf{f}_{v,i}^{\mathrm{proj}} \cdot \mathbf{t}_j}{\tau},
\label{eq:sim_matrix}
\end{equation}
where $\tau$ is a learnable temperature parameter. All projected visual and text embeddings are $\ell_2$-normalized before similarity computation.

The image-to-text and text-to-image cross-modal cross-entropy losses are defined as:
\begin{equation}
\mathcal{L}_{\mathrm{CCE}}(\mathbf{f}_v^{\mathrm{proj}}, \mathbf{t}) =
\frac{1}{B}\sum_{i=1}^{B}
- \log
\frac{\exp(S_{ii})}{\sum_{j=1}^{B} \exp(S_{ij})},
\label{eq:l_i2t}
\end{equation}

\begin{equation}
\mathcal{L}_{\mathrm{CCE}}(\mathbf{t}, \mathbf{f}_v^{\mathrm{proj}}) =
\frac{1}{B}\sum_{i=1}^{B}
- \log
\frac{\exp(S_{ii})}{\sum_{j=1}^{B} \exp(S_{ji})}.
\label{eq:l_t2i}
\end{equation}

The symmetric cross-modal alignment loss is
\begin{equation}
\mathcal{L}_{\mathrm{cm}} =
\frac{1}{2}
\left(
\mathcal{L}_{\mathrm{CCE}}(\mathbf{f}_v^{\mathrm{proj}}, \mathbf{t})
+
\mathcal{L}_{\mathrm{CCE}}(\mathbf{t}, \mathbf{f}_v^{\mathrm{proj}})
\right).
\label{eq:l_cm}
\end{equation}

This bidirectional formulation is particularly important in longitudinal settings (as also supported by the ablation results), where identity representations may drift over time. Image-to-text alignment alone constrains visual embeddings toward identity-conditioned text anchors, but does not enforce reciprocal consistency. By optimizing both directions, symmetric alignment provides additional regularization across modalities, providing additional regularization that improves robustness under temporal shift.

Batch-hard triplet losses are applied independently to the visual and projected embeddings:
\begin{equation}
\mathcal{L}_{\mathrm{tri}} =
\mathcal{L}_{\mathrm{tri}}^{(v)}(\mathbf{f}_v, y)
+
\mathcal{L}_{\mathrm{tri}}^{(p)}(\mathbf{f}_v^{\mathrm{proj}}, y).
\label{eq:l_tri}
\end{equation}

Triplet losses are computed independently within each embedding space: anchors, positives, and negatives are formed exclusively from either $\mathbf{f}_v$ or $\mathbf{f}_v^{\mathrm{proj}}$, and no cross-space triplets are used. All embeddings are $\ell_2$-normalized prior to distance computation, and the Euclidean distance is employed for both triplet optimization and retrieval.

Auxiliary classification losses are applied to the logits $\mathbf{s}_v$ and $\mathbf{s}_p$:
\begin{equation}
\mathcal{L}_{\mathrm{aux}} =
\mathcal{L}_{\mathrm{CE}}(\mathbf{s}_v, y)
+
\mathcal{L}_{\mathrm{CE}}(\mathbf{s}_p, y).
\label{eq:l_aux}
\end{equation}

The overall training objective is
\begin{equation}
\mathcal{L} =
\lambda_{\mathrm{tri}} \mathcal{L}_{\mathrm{tri}}
+
\lambda_{\mathrm{cm}} \mathcal{L}_{\mathrm{cm}}
+
\mathcal{L}_{\mathrm{aux}}.
\label{eq:overall_loss}
\end{equation}

Although $\mathbf{f}_v^{\mathrm{proj}}$ is a linear projection of $\mathbf{f}_v$, supervising both spaces serves distinct functions. The two representations reside in different dimensional spaces and are used jointly at inference through feature concatenation. Applying metric supervision in both spaces encourages consistent identity separation before and after projection and ensures that LoRA-induced updates remain discriminative under the frozen CLIP projection.

During training, only the LoRA parameters, learnable prompt context tokens, metadata-related parameters, auxiliary identity heads, and batch normalization layers are optimized, while all pretrained CLIP backbone parameters remain frozen.

\subsection{Inference}
\label{subsec:inference}
In this study, ReID is formulated as retrieval rather than classification, allowing the model to generalize to unseen identities without requiring a fixed label space at inference.

At inference time, only the LoRA-adapted vision encoder is retained, and all text-related components and auxiliary supervision heads are removed. Given a query image and a gallery set, ReID is performed by nearest-neighbor retrieval using the Euclidean distance between fused visual descriptors. Specifically, we extract the visual embedding $\mathbf{f}_v$ and its CLIP-projected counterpart $\mathbf{f}_v^{\mathrm{proj}}$, concatenate their $\ell_2$-normalized representations, and apply $\ell_2$ normalization to the concatenated vector to obtain the final descriptor. the Euclidean distance between the resulting descriptors is used to rank the gallery images.

Although triplet supervision is applied independently to $\mathbf{f}_v$ and $\mathbf{f}_v^{\mathrm{proj}}$ during training, both embeddings are optimized for identity discrimination. Their concatenation at inference aggregates complementary information from the raw visual and CLIP-aligned spaces without introducing additional trainable parameters. The use of concatenation is motivated by the observation that the visual and projected representations capture complementary identity information. This is supported by the experimental results, where supervision in both embedding spaces yields substantially higher performance than relying on either representation individually. Consequently, concatenation provides a simple parameter-free mechanism for preserving information from both embedding spaces without introducing additional trainable fusion modules.

\begin{figure*}[!htbp]
    \centering
    \includegraphics[width=0.7\textwidth]{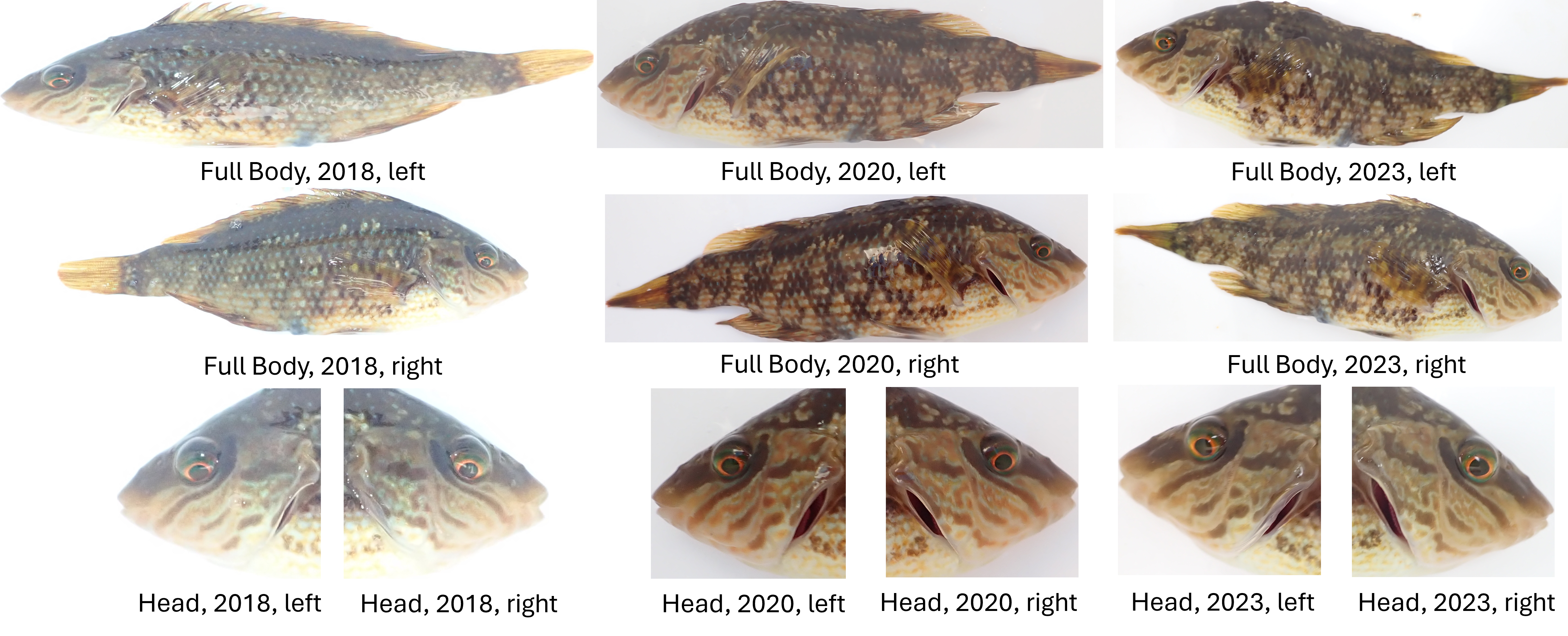}
    \caption{Multiple observations of the same individual from the Melops dataset~\cite{sordalen2025melopsreid} across different capture years. The top and middle rows show full-body lateral views from different time points and body sides, illustrating natural growth, size variation, and side-specific appearance differences. The bottom row presents corresponding head crops, highlighting fine-grained facial patterns that remain discriminative despite longitudinal morphological and coloration changes, while also exhibiting the left-right asymmetry.}
    \label{fig:dataset}
\end{figure*}

\section{Experimental Analysis}
\label{sec:experiments}
In this section, we first describe the dataset, evaluation protocols, and implementation details. We then present quantitative results under different evaluation settings, followed by ablation studies that analyze the contribution of individual components of the proposed framework. Finally, we report additional experiments on external benchmarks to assess generalization.

\subsection{Melops Dataset}
\label{subsec:melops}

We evaluated our approach in the Melops dataset~\cite{sordalen2025melopsreid,Sordalen2026wild}, a large-scale longitudinal image dataset for individual ReID of wild corkwing wrasse (\textit{Symphodus melops}). The dataset comprises 24578 images of 9861 PIT-tagged individuals collected over a seven-year capture–mark–recapture program (2018–2024) in Western Norway. Each fish was photographed from both left and right sides under standardized imaging conditions against a white background with a color reference card, enabling consistent cropping and color normalization.

A defining characteristic of Melops is its temporal depth. Of the 9861 individuals, 1883 were recaptured at least once, yielding 2916 resighting events and 8524 images of repeated encounters. Capture intervals range from within-season recaptures to multi-year spans, reflecting natural growth, seasonal coloration changes, and morphological variation. At the same time, the majority of individuals were observed only once, creating a highly imbalanced identity distribution that poses a significant challenge for metric-based learning. 

In addition to temporal variation, \textit{Symphodus melops} exhibits pronounced left–right asymmetry in pigmentation and patterning. As illustrated in Figure~\ref{fig:dataset}, longitudinal captures of the same individual show substantial variation in size, coloration, and side-specific appearance, with distinct visual patterns often present on each lateral side. Such an asymmetry increases intra-individual variability and prevents trivial memorization of the mirrored views. Furthermore, biological studies suggest that head and facial regions contain stable and identity-relevant morphological cues in fish, particularly around the operculum, eye, and snout regions~\cite{ellis_visual_2026}. This motivates the examination of head-based crops in addition to full-body views.

Furthermore, this dataset provides rich capture-level metadata, including total body length, sex, capture location, and precise capture date recorded in day-sequence format. 
These attributes reflect biologically meaningful factors such as growth, reproductive state, and seasonal variation, which directly influence visual appearance over time. In our experiments, we use both the full-body and head crop variants provided with the dataset. Temporal information is derived from the normalized day-sequence index to represent absolute progression across the study period, enabling systematic evaluation under time-aware protocols. Besides, the body length and capture year are also incorporated as numerical attributes into our model.

\subsection{Evaluation Protocols}
\label{subsec:protocol}
To comprehensively evaluate generalization under different deployment scenarios, we define four evaluation protocols that vary in identity overlap and temporal ordering. Table~\ref{tab:dataset_splits} summarizes the resulting data splits under each protocol.

\noindent \textbf{Closed-Set (CS).} 
In the CS protocol, identities appearing in the test split overlap with those in the training split, i.e., 
\(\mathcal{I}_{\text{test}} \subseteq \mathcal{I}_{\text{train}}\). 
The dataset is partitioned at the image level within each identity such that training and testing images are disjoint. 
Approximately 70\% of the images per identity are allocated to training, with the remainder used for evaluation and divided into query and gallery sets. 
Query and gallery images are image-disjoint but identity-overlapping, following standard closed-set ReID practice~\cite{li2020atrw,li2023clip,wu2021deep,jiao2023toward,cermak2024wildfusion,adam2024seaturtleid}. 
For identities with fewer than three samples, all samples are assigned to the training split. \\

\noindent \textbf{Open-Set (OS).} 
Following the protocol in~\cite{adam2024seaturtleid}, the test set contains a mixture of seen and unseen identities. 
Seen identities satisfy \(\mathcal{I}_{\text{test}}^{\text{seen}} \subseteq \mathcal{I}_{\text{train}}\), 
while unseen identities satisfy \(\mathcal{I}_{\text{test}}^{\text{unseen}} \cap \mathcal{I}_{\text{train}} = \emptyset\), 
with \(\mathcal{I}_{\text{test}} = \mathcal{I}_{\text{test}}^{\text{seen}} \cup \mathcal{I}_{\text{test}}^{\text{unseen}}\) and 
\(\mathcal{I}_{\text{test}}^{\text{seen}} \cap \mathcal{I}_{\text{test}}^{\text{unseen}} = \emptyset\). 
Within the test set, 20\% of identities are designated as unseen, and the remaining 80\% as seen identities overlapping with \(\mathcal{I}_{\text{train}}\). 
Test identities are distributed across both query and gallery splits, ensuring that query and gallery share the same identity set.\\

\noindent \textbf{Time-Aware Closed-Set (TACS).} 
We construct a chronologically constrained CS protocol in which identities overlap between training and testing, but test images are restricted to the final year of data collection, while all earlier years are used for training. 
This enforces strict temporal separation and evaluates forward-time generalization under realistic longitudinal conditions. \\

\noindent \textbf{Time-Aware Open-Set (TAOS).}
TAOS follows the same identity partitioning as OS, with the test set comprising both seen and unseen identities. 
However, evaluation images are restricted exclusively to observations from the final year, while all earlier years are used for training. 
This protocol evaluates generalization across both identity novelty and longitudinal appearance changes under forward-time conditions.

\begin{table}[t]
\centering
\caption{Statistics of the defined evaluation protocols. 
Identity counts refer to the unique identities per split.}
\label{tab:dataset_splits}
\resizebox{0.45\textwidth}{!}{%
\begin{tabular}{lccccc}
\toprule
\textbf{Protocol}  & \textbf{Train IDs} & \textbf{Train Imgs} & \textbf{Query IDs} & \textbf{Query Imgs} & \textbf{Gallery Imgs} \\
\midrule
\rowcolor{lightblue}
CS    & 9814 & 20656 & 1837 & 1843 & 1953 \\
OS    & 9470 & 19761 & 1872 & 2254 & 2533 \\
\rowcolor{lightblue}
TACS  & 9290 & 22420 & 37   & 76   & 84   \\
TAOS  & 9209 & 22225 & 86   & 173  & 188  \\
\bottomrule
\end{tabular}
}
\end{table}

\subsection{Evaluation Metrics}
\label{subsec:metrics}

We employ standard ranking-based metrics for ReID evaluation. 
Let \( f : \mathcal{X} \rightarrow \mathbb{R}^d \) denote the embedding function. 
Gallery samples are ranked for each query according to ascending Euclidean distance \( \| f(q) - f(g) \|_2 \). \\

\noindent \textbf{Cumulative Matching Characteristic (CMC).}
The CMC curve measures the probability that at least one correct match appears within the top-\(k\) ranked gallery samples. 
We report Rank-1 and Rank-5 accuracy, defined as
\begin{equation}
\text{Rank-}k 
=
\frac{1}{|\mathcal{Q}|}
\sum_{q \in \mathcal{Q}}
\mathbb{1}
\left[
\exists\, g \in \mathcal{G}_q^{+}
:\;
\text{rank}(g) \leq k
\right],
\label{eq:cmc}
\end{equation}
where \(\mathcal{Q}\) denotes the query set, 
\(\mathcal{G}_q^{+}\) is the set of gallery samples sharing the same identity as query \(q\). \\

\noindent \textbf{Mean Average Precision (mAP).}
To account for multiple correct matches per query, we compute mean Average Precision (mAP) as
\[
\text{mAP}
=
\frac{1}{|\mathcal{Q}|}
\sum_{q \in \mathcal{Q}}
\text{AP}(q),
\]
where the Average Precision for query \(q\) is
\begin{equation}
\text{AP}(q)
=
\frac{1}{|\mathcal{G}_q^{+}|}
\sum_{k: g_k \in \mathcal{G}_q^{+}}
P_q(k),
\label{eq:ap}
\end{equation}
and
\begin{equation}
P_q(k) =
\frac{\left| \left\{ g_j \in \mathcal{G}_q^{+} \;\middle|\; j \leq k \right\} \right|}{k}
\label{eq:precision_at_k}
\end{equation}
denotes the precision at rank \(k\).

\subsection{Implementation Details}
\label{subsec:implementation}

\noindent \textbf{Training Setup.}
Our model and all ablation variants are trained under identical optimization settings. We use a CLIP ViT-B/16 backbone~\cite{radford2021learning} initialized from publicly available pretrained weights to remain consistent with prior CLIP-based ReID approaches~\cite{li2023clip,wu2024individual} and to emphasize parameter efficiency.
The vision encoder has feature dimension $d_v = 768$ and is projected into the shared embedding space of dimension $d_t = 512$. 
For the Melops dataset, we apply standard ReID augmentations, including random horizontal flip, padding, random crop, normalization, and random erasing~\cite{li2023clip,wu2021deep}. 
Unless otherwise specified, backbone parameters remain frozen, and only task-specific adaptation modules are optimized.

Training is performed using a composite objective consisting of two batch-hard triplet losses, symmetric cross-modal contrastive cross-entropy losses, and two auxiliary identity classification losses. 
The triplet losses are computed independently in the visual feature space and the CLIP-projected feature space. 
All embeddings are $\ell_2$-normalized before distance computation for triplet optimization and retrieval. 
Projected visual and text embeddings are $\ell_2$-normalized before similarity computation for the cross-modal loss.

In line with prior art~\cite{li2023clip,wu2024individual}, the triplet loss weight $\lambda_{\mathrm{tri}}$ is set to 1, the cross-modal loss weight $\lambda_{\mathrm{cm}}$ is set to 1, and the temperature parameter $\tau$ in the similarity computation is set to 1.

The batch composition follows a standard PK sampling strategy with $P = 16$ identities and $K = 4$ images per identity per mini-batch (64 images in total). 
Batch-hard triplets are constructed within each mini-batch following the standard batch-hard mining strategy. 

We optimize the model using AdamW~\cite{loshchilov2017decoupled} with a learning rate of $5 \times 10^{-4}$ and the weight decay of $1 \times 10^{-5}$ in line with~\cite{li2023clip}. 
Training is conducted for up to 60 epochs with a batch size of 64. 
All experiments were performed on a single NVIDIA GeForce RTX 5090 GPU. \\

\noindent \textbf{Our Method.}
Unless otherwise specified, we use a LoRA rank of $r = 16$ and set the LoRA scaling factor $\alpha = 16$. 
LoRA is applied to the query, key, and value linear projection layers within each multi-head self-attention block. 
LoRA parameters are initialized such that the adaptation branch is inactive at the beginning of training (i.e., initialized to produce zero effective update), ensuring that the initial forward pass recovers the pretrained CLIP behavior.

The prompt consists of fixed prefix tokens, a fixed class token, a fixed identity token, and a set of learnable context tokens. 
Learnable context tokens are randomly initialized, and their number is set to $M = 4$ per identity. 
To reduce sensitivity to prompt structure, we employ prompt-order ensembling with three layouts (front, middle, and end)~\cite{zhou2022coop}. 
For the middle layout, the context tokens are split evenly into two halves. 
The final text representation is obtained by averaging the $\ell_2$-normalized embeddings produced by the frozen CLIP text encoder across all prompt layouts.

Each identity is assigned a unique, fixed identity token initialized randomly in the CLIP text embedding space. 
All images belonging to the same identity share the same identity token, while different identities are assigned distinct tokens. 
Identity tokens are not learnable and remain fixed throughout training. 
Both identity and class tokens are discarded at inference time.

Metadata is image-specific and consists of numerical attributes such as capture time and physical measurements. 
All scalar metadata attributes are normalized using per-attribute min-max scaling computed on the training set, and the same scaling parameters are reused during testing to prevent information leakage. During training, metadata embeddings are appended to the end of the prompt sequence and are not used at inference time, ensuring that the deployed model remains purely visual.

For sinusoidal metadata encoding, fixed exponentially spaced frequencies are used following the standard sine-cosine positional encoding formulation in Transformers. 
Encodings are computed independently for each metadata dimension and concatenated before linear projection. We use 16 frequencies per metadata dimension, following common practice in sinusoidal positional encoding, resulting in 32 encoded features (sine and cosine pairs) per attribute and a total pre-projection dimension of $d_{\mathrm{PE}} = 32\,n_a$.

For FiLM-based conditioning, learnable attribute basis tokens $\mathbf{T}_a \in \mathbb{R}^{M_a \times d_t}$ are shared across samples and optimized during training. 
Scaling and shifting parameters are generated from metadata via a learnable projection and applied feature-wise to the attribute basis tokens. 
The number of attribute basis tokens is set to $M_a = n_a$, corresponding to one basis token per metadata dimension. 
All metadata-related projection layers and FiLM parameters are trained jointly with the remaining learnable components.

Only the LoRA parameters in the vision encoder, learnable context tokens, metadata-related parameters (including sinusoidal projection or FiLM parameters), auxiliary identity classification heads, and batch normalization layers are updated during training. 
All remaining CLIP backbone parameters remain frozen. 
At inference time, only the LoRA-adapted vision encoder is retained. 
All text-related components, metadata embeddings, and auxiliary supervision heads are removed. 
ReID is performed by nearest-neighbor retrieval using Euclidean distance between $\ell_2$-normalized visual descriptors obtained from the concatenated embeddings. \\

\noindent \textbf{Alternative Visual Adaptation Methods.}
To evaluate the effectiveness of LoRA-based adaptation, we compare against several alternative lightweight visual adapters. 
All alternative modules are inserted into the frozen CLIP vision encoder and trained using the same composite objective and optimization setup described above. \\

\paragraph{Linear Head.}
As a minimal baseline, no intermediate adaptation modules are introduced. 
Frozen CLIP visual features $\mathbf{f}_v \in \mathbb{R}^{d_v}$ are first batch-normalized and then projected directly to the identity classification space via a single linear layer,
$\mathbf{s} = \mathbf{W}_c\,\mathrm{BN}(\mathbf{f}_v)$,
where $\mathbf{W}_c \in \mathbb{R}^{C \times d_v}$ is a trainable weight matrix. 
This baseline is trained using the same auxiliary cross-entropy loss as in our method. \\

\paragraph{Bottleneck Adapter.}
A residual bottleneck adapter is inserted after each transformer block of the CLIP vision encoder. 
The adapter projects the feature $\mathbf{f} \in \mathbb{R}^{d_v}$ into a lower-dimensional bottleneck space of dimension $b$, and then maps it back to the original dimension:
$\mathbf{h}_b = \sigma(\mathbf{W}_{\mathrm{down}} \mathbf{f})$, 
$\mathbf{f}' = \alpha\, \sigma(\mathbf{W}_{\mathrm{up}} \mathbf{h}_b) + (1-\alpha)\mathbf{f}$,
where $\mathbf{W}_{\mathrm{down}} \in \mathbb{R}^{b \times d_v}$ and 
$\mathbf{W}_{\mathrm{up}} \in \mathbb{R}^{d_v \times b}$ are learnable projection matrices, 
$\sigma(\cdot)$ denotes LeakyReLU activation, and $\alpha$ controls residual scaling. \\

\paragraph{1D Convolutional Adapter.}
A 1D convolution is applied across the token sequence before bottleneck projection. 
Specifically,
$\mathbf{f}_{\mathrm{conv}} = \mathrm{Conv1D}(\mathbf{f}; k=1)$ and 
$\mathbf{h}_b = \sigma(\mathbf{W}_{\mathrm{down}}\, \mathrm{IN}(\mathbf{f}_{\mathrm{conv}}))$, 
where $\mathrm{Conv1D}(\cdot; k=1)$ denotes channel-wise convolution with kernel size $k=1$, 
$\mathrm{IN}(\cdot)$ applies instance normalization, and $\sigma(\cdot)$ is LeakyReLU. 
Note that kernel size $k=1$ performs channel mixing without expanding the receptive field across neighboring tokens. \\

\paragraph{2D Convolutional Adapter.}
Patch tokens (excluding the class token) are reshaped into a spatial grid and processed with a 2D convolution before bottleneck projection. 
Formally,
$\mathbf{f}_{2D} = \mathrm{Reshape}(\mathbf{f}_{1:N}; H \times W)$ and 
$\mathbf{f}_{\mathrm{conv}} = \mathrm{Conv2D}(\mathbf{f}_{2D}; k=3, p=1)$, 
where $\mathbf{f}_{1:N}$ denotes the patch tokens, $H = W = \sqrt{N}$ for square grids, and $(k,p)$ denote kernel size and padding. 
The convolved features are flattened, concatenated with the class token, and passed through the bottleneck adapter. \\

\paragraph{Transformer Adapter.}
A lightweight transformer encoder layer with multi-head self-attention is inserted before bottleneck projection to model inter-token dependencies. 
Formally,
$\mathbf{f}_{\mathrm{attn}} = \mathrm{TransformerLayer}(\mathbf{f}; n_{\mathrm{heads}}=8)$, 
where $n_{\mathrm{heads}}=8$ denotes the number of attention heads. 
The resulting features are then passed through the bottleneck projection:
$\mathbf{h}_b = \mathbf{W}_{\mathrm{down}} \mathbf{f}_{\mathrm{attn}}$. \\

We evaluate inserting adapters after different transformer blocks and observe that inserting adapters after all transformer blocks yields the best performance. 
Each transformer block employs an independent set of adapter parameters, and parameters are not shared across layers. 
The bottleneck dimension is set to $b = 512$. 
Adapter weights are initialized using Kaiming initialization~\cite{he2015delving}, and the initial adapter contribution is scaled to avoid disrupting pretrained CLIP representations at the beginning of training. 
LeakyReLU is used as the activation function in bottleneck projections. 
No additional dropout or normalization layers are introduced beyond those explicitly described above. \\

\noindent \textbf{Other CLIP-based ReID Methods.}
We compare our method against CLIP-ReID~\cite{li2023clip} and IndivAID~\cite{wu2024individual}, using their official public implementations. For both methods, only dataset-specific configurations (paths and dataset loaders) are adapted to match our evaluation benchmarks. The original architectures, loss functions, optimizer types, learning rate schedules, sampling strategies, data augmentation, and evaluation protocols strictly follow the released repository defaults.

All methods share the same CLIP backbone and follow closely related training paradigms, resulting in comparable optimization settings. Each method is evaluated under its original configuration, while using identical dataset splits and evaluation metrics for consistency. Following common benchmarking practice, the compared methods retain their published training recipes and hyperparameter settings. Training is run until convergence when necessary, and the best-performing checkpoint is selected for reporting.

\subsection{Results}
\label{subsec:results}

This section reports the experimental evaluation of the proposed framework. We first present results on the Melops dataset under standard CS and OS protocols for both full-body and head-crop inputs. We then evaluate longitudinal robustness using TACS and TAOS settings that enforce temporal separation between training and testing data (Sec.~\ref{subsec:sota}). Next, we analyze the effects of continuous metadata conditioning (Sec.~\ref{subsec:metaCond}), followed by a series of ablation studies examining the contributions of low-rank visual adaptation (Sec.~\ref{subsec:abl1}), prompt learning and prompt-order ensembling (Sec.~\ref{subsec:abl2}), identity tokens (Sec.~\ref{subsec:abl3}), auxiliary supervision (Sec.~\ref{subsec:abl4}), cross-modal alignment (Sec.~\ref{subsec:abl5}), and triplet learning strategies (Sec.~\ref{subsec:abl6}). We further analyze identification performance as a function of temporal distance (Sec.~\ref{subsec:temporal}) and present an analysis of temporal and body-side effects (Sec.~\ref{subsec:stat}). Finally, we report efficiency comparisons (Sec.~\ref{subsec:pea}) and assess generalization on additional animal ReID benchmarks (Sec.~\ref{subsec:others}).

\begin{table}[tb!]
\centering
\caption{Closed-set performance comparison. Results are reported as mean $\pm$ standard deviation over five random seeds. Bold indicates the best result. Full fine-tuning (FT) refers to jointly optimizing both the vision and text encoders of CLIP-B/16.
}
\label{tab:sotaCLOSED}
\resizebox{0.5\textwidth}{!}{%
\begin{tabular}{lcccc}
\toprule
\textbf{Method} & \textbf{Crop} & \textbf{mAP} & \textbf{Rank-1} & \textbf{Rank-5} \\
\midrule
CLIP-B/16 (Full FT)~\cite{radford2021learning} & Head & 52.70 $\pm$ 0.35 & 45.80 $\pm$ 0.53 & 62.20 $\pm$ 0.30 \\
CLIP-B/16 (Full FT)~\cite{radford2021learning} & Body & 52.10 $\pm$ 0.63 & 44.50 $\pm$ 0.80 & 61.70 $\pm$ 0.70 \\
\midrule
CLIP-ReID~\cite{li2023clip} & Head & 44.32 $\pm$ 0.31 & 37.58 $\pm$ 0.46 & 52.90 $\pm$ 0.20 \\
CLIP-ReID~\cite{li2023clip} & Body & 44.34 $\pm$ 0.22 & 37.92 $\pm$ 0.19 & 52.36 $\pm$ 0.61 \\
\midrule
IndivAID~\cite{wu2024individual} & Head & 43.16 $\pm$ 0.52 & 36.30 $\pm$ 0.53 & 52.12 $\pm$ 0.57 \\
IndivAID~\cite{wu2024individual} & Body & 44.02 $\pm$ 0.43 & 37.59 $\pm$ 0.62 & 52.60 $\pm$ 0.56 \\
\midrule
\rowcolor{lightblue}
Ours & Head & 52.48 $\pm$ 0.33 & \textbf{45.04 $\pm$ 0.47} & 62.38 $\pm$ 0.28 \\
\rowcolor{lightblue}
Ours & Body & \textbf{53.00 $\pm$ 0.70} & 44.80 $\pm$ 0.75 & \textbf{64.32 $\pm$ 0.62} \\
\bottomrule
\end{tabular}}
\end{table}

\subsubsection{Comparison with Prior Art}
\label{subsec:sota}

\paragraph{Closed-Set Evaluation.} 
Table~\ref{tab:sotaCLOSED} reports performance comparisons under the CS protocol. Our approach consistently outperforms CLIP-ReID~\cite{li2023clip} and IndivAID~\cite{wu2024individual} by a substantial margin in both head and body crop settings. In addition, we compare against full fine-tuning of the same CLIP-B/16 backbone. On body crops, our method improves over CLIP-ReID by +8.66 points and over IndivAID by +8.98 points. 
On head crops, we exceed both baselines by more than 8 mAP points. Improvements are also reflected in Rank-1 and Rank-5 accuracy, where our method consistently achieves the highest scores. Compared to full fine-tuning, our method achieves comparable performance on head crops (52.48 vs. 52.70 mAP) and slightly improves performance on body crops (53.00 vs. 52.10 mAP), while requiring substantially fewer trainable parameters. Across methods, body crops yield slightly stronger performance than head crops, particularly for our approach (+0.52 mAP). This suggests that global morphological cues complement the distinctive facial patterns characteristic of the species, and that leveraging full-body information can provide additional discriminative signal under closed-set conditions.
These improvements are consistent with the architectural differences between the methods. In contrast to CLIP-ReID, which applies staged prompt optimization and full fine-tuning strategies, our method jointly optimizes low-rank visual adaptation and learnable prompts while keeping the backbone frozen. Compared to IndivAID, which employs a multi-stage pipeline with identity-specific token constructs, our approach integrates identity-aware prompt learning and visual adaptation within a unified training framework. 
Overall, these results indicate that the proposed parameter-efficient adaptation achieves a strong balance between performance and efficiency, matching or exceeding full fine-tuning while avoiding the need to update the entire backbone.
Empirically, this parameter-efficient adaptation strategy yields higher closed-set performance and stable results across random seeds.

\begin{table}[tb!]
\centering
\caption{Open-set performance comparison. Results are reported as mean $\pm$ standard deviation over five random seeds. Bold indicates the best result.}
\label{tab:sotaOPEN}
\resizebox{0.5\textwidth}{!}{%
\begin{tabular}{lcccc}
\toprule
\textbf{Method} & \textbf{Crop} & \textbf{mAP} & \textbf{Rank-1} & \textbf{Rank-5} \\
\midrule
CLIP-ReID~\cite{li2023clip} & Head & 32.40 $\pm$ 0.52 & 27.40 $\pm$ 0.53 & 42.60 $\pm$ 0.57 \\
CLIP-ReID~\cite{li2023clip} & Body & 30.90 $\pm$ 0.20 & 26.00 $\pm$ 0.36 & 39.00 $\pm$ 0.41 \\
\midrule
IndivAID~\cite{wu2024individual} & Head & 29.41 $\pm$ 0.40 & 24.45 $\pm$ 0.37 & 38.02 $\pm$ 0.45 \\
IndivAID~\cite{wu2024individual} & Body & 30.05 $\pm$ 0.58 & 25.11 $\pm$ 0.24 & 38.91 $\pm$ 0.21 \\
\midrule
\rowcolor{lightblue}
Ours & Head & \textbf{38.30 $\pm$ 0.33} & \textbf{32.60 $\pm$ 0.30} & \textbf{48.60 $\pm$ 0.17} \\
\rowcolor{lightblue}
Ours & Body & 37.90 $\pm$ 0.22 & 31.60 $\pm$ 0.28 & 48.20 $\pm$ 0.13 \\
\bottomrule
\end{tabular}}
\end{table}

\paragraph{Open-Set Evaluation.}
Under the OS protocol (Table~\ref{tab:sotaOPEN}), performance decreases for all methods compared to the CS protocol, reflecting the increased difficulty of generalizing to unseen identities. 
Nevertheless, our method maintains a substantial performance margin over prior work. On head crops, we improve over CLIP-ReID by +5.90 points and over IndivAID by +8.89 points. 
On body crops, we surpass both baselines by more than 7 mAP points.
Improvements are consistent across Rank-1 and Rank-5 accuracy. 
Notably, in contrast to the closed-set setting where body crops slightly outperform head crops, the head crop yields the strongest performance under open-set conditions. This suggests that localized facial patterns may generalize more robustly across unseen identities, whereas global body cues may be more sensitive to inter-individual variability. Overall, these results indicate that our joint prompt and low-rank visual adaptation strategy generalizes more effectively to novel identities than prior CLIP-based approaches.

\paragraph{Time-aware Evaluation.}
We further evaluate all methods under temporally structured protocols, where training samples precede test samples in time. 

\textit{Time-aware Closed-set.} 
Table~\ref{tab:timeaware_closed} evaluates robustness to temporal appearance changes within known individuals.
Across both head and body crops, our method outperforms CLIP-ReID and IndivAID. 
Without metadata conditioning, the proposed model exceeds CLIP-ReID by +2.00 and +1.90 mAP points, respectively. When metadata conditioning is introduced (Ours + Meta), performance further improves +2.90 mAP on head crops and +1.30 mAP on body crops. Improvements are also reflected in Rank-1 and Rank-5 accuracy, where Ours + Meta consistently achieves the strongest performance.
These results suggest that incorporating continuous temporal metadata enhances robustness to appearance variations across capture periods. 
Since prior methods do not exploit explicit metadata conditioning, the observed gains indicate the benefit of integrating temporal cues into prompt-based adaptation under temporally constrained evaluation.

\begin{table}[tb!]
\centering
\caption{Time-aware closed-set performance comparison. 
Results are reported as mean $\pm$ standard deviation over five random seeds. 
Bold indicates the best result in each column.}
\label{tab:timeaware_closed}
\resizebox{0.5\textwidth}{!}{%
\begin{tabular}{lcccc}
\toprule
\textbf{Method} & \textbf{Crop} & \textbf{mAP} & \textbf{Rank-1} & \textbf{Rank-5} \\
\midrule
CLIP-ReID~\cite{li2023clip} & Head & 61.00 $\pm$ 0.18 & 54.90 $\pm$ 0.16 & 80.20 $\pm$ 0.19 \\
CLIP-ReID~\cite{li2023clip} & Body & 59.40 $\pm$ 0.14 & 54.60 $\pm$ 0.17 & 76.30 $\pm$ 0.15 \\
\midrule
IndivAID~\cite{wu2024individual} & Head & 59.18 $\pm$ 0.22 & 55.26 $\pm$ 0.24 & 81.89 $\pm$ 0.20 \\
IndivAID~\cite{wu2024individual} & Body & 57.06 $\pm$ 0.21 & 53.95 $\pm$ 0.23 & 75.00 $\pm$ 0.18 \\
\midrule
\rowcolor{lightblue}
Ours & Head & 63.00 $\pm$ 0.11 & 57.90 $\pm$ 0.17 & 82.90 $\pm$ 0.09 \\
\rowcolor{lightblue}
Ours & Body & 61.30 $\pm$ 0.09 & 52.60 $\pm$ 0.12 & 81.60 $\pm$ 0.13 \\
\rowcolor{lightblue}
Ours + Meta (Year+FiLM) & Head & \textbf{65.90 $\pm$ 0.12} & \textbf{60.50 $\pm$ 0.15} & \textbf{84.20 $\pm$ 0.11} \\
\rowcolor{lightblue}
Ours + Meta (Year+FiLM) & Body & \textbf{62.60 $\pm$ 0.10} & \textbf{54.94 $\pm$ 0.11} & \textbf{81.63 $\pm$ 0.17} \\

\bottomrule
\end{tabular}}
\end{table}

\textit{Time-aware Open-set.}
Table~\ref{tab:timeaware_open} evaluates the ability of each method to generalize to unseen individuals under temporal distribution shifts.
Compared to prior methods, our approach consistently achieves higher performance across both crop types. 
Without metadata conditioning, our model already improves over CLIP-ReID by +3.50 mAP on head crops and +0.70 mAP on body crops. When metadata conditioning is enabled (Ours + Meta), performance further increases. 
The benefit of metadata conditioning is particularly pronounced for body crops, where mAP improves by +2.52 points relative to the base and by +3.22 points over CLIP-ReID. 
Rank-1 accuracy shows a similar trend, increasing from 35.30\% to 39.94\% for body crops when metadata is incorporated. For head crops, metadata provides a more modest but consistent gain (+0.62 mAP), indicating that facial patterns may already encode stronger identity cues under temporal shifts. Overall, these results demonstrate that combining prompt learning, low-rank visual adaptation, and continuous metadata conditioning enhances robustness in the presence of both unseen identities and temporal appearance changes. We note that alternative metadata configurations can yield higher performance for specific crops, as shown in Table~\ref{tab:timeaware_open_ablation}. However, for clarity and consistency, we report a single configuration across both crops in this table.

\begin{table}[tb!]
\centering
\caption{Time-aware open-set performance comparison. 
Results are reported as mean $\pm$ standard deviation over five random seeds. 
Bold indicates the best result in each column.}
\label{tab:timeaware_open}
\resizebox{0.5\textwidth}{!}{%
\begin{tabular}{lcccc}
\toprule
\textbf{Method} & \textbf{Crop} & \textbf{mAP} & \textbf{Rank-1} & \textbf{Rank-5} \\
\midrule
CLIP-ReID~\cite{li2023clip} & Head & 47.90 $\pm$ 0.17 & 42.80 $\pm$ 0.18 & 67.60 $\pm$ 0.19 \\
CLIP-ReID~\cite{li2023clip} & Body & 42.20 $\pm$ 0.15 & 35.30 $\pm$ 0.16 & 54.30 $\pm$ 0.14 \\
\midrule
IndivAID~\cite{wu2024individual} & Head & 40.63 $\pm$ 0.23 & 33.53 $\pm$ 0.22 & 59.54 $\pm$ 0.21 \\
IndivAID~\cite{wu2024individual} & Body & 38.82 $\pm$ 0.24 & 31.79 $\pm$ 0.20 & 54.34 $\pm$ 0.19 \\
\midrule
\rowcolor{lightblue}
Ours & Head & 51.40 $\pm$ 0.12 & 43.40 $\pm$ 0.09 & 67.10 $\pm$ 0.07 \\
\rowcolor{lightblue}
Ours & Body & 42.90 $\pm$ 0.11 & 35.30 $\pm$ 0.14 & 55.50 $\pm$ 0.11 \\
\rowcolor{lightblue}
Ours + Meta (DaySeq+Sin) & Head & \textbf{52.02 $\pm$ 0.12} & \textbf{45.74 $\pm$ 0.09} & \textbf{67.14 $\pm$ 0.10} \\
\rowcolor{lightblue}
Ours + Meta (DaySeq+Sin) & Body & \textbf{45.42 $\pm$ 0.15} & \textbf{39.94 $\pm$ 0.13} & \textbf{59.54 $\pm$ 0.11} \\
\bottomrule
\end{tabular}}
\end{table}

\subsubsection{Effect of Metadata Conditioning}
\label{subsec:metaCond}

We investigate the impact of different metadata conditioning strategies relative to the no-metadata baseline under both the time-aware closed-set (Table~\ref{tab:timeaware_close_ablation}) and time-aware open-set (Table~\ref{tab:timeaware_open_ablation}) protocols. In addition to our continuous metadata conditioning, we examine a discretized textual encoding strategy inspired by~\cite{li2025metawild}. In this variant, numerical attributes are first partitioned into a fixed number of bins defined on the whole dataset. Each bin is mapped to a descriptive textual token (e.g., ``small'', ``medium'', ``large'' for body length; ``early'', ``mid'', ``late'' for temporal progression; and calendar year descriptors for capture year). The resulting text embedding is appended to the prompt representation in place of the continuous metadata embedding. Apart from this metadata conversion step, all architectural components, optimization procedures, and inference settings remain unchanged.

Table~\ref{tab:timeaware_close_ablation} shows that the discretized textual variant does not consistently improve performance. This suggests that coarse symbolic conditioning may introduce artificial discontinuities in the embedding space, which can negatively affect fine-grained identity discrimination, particularly for head crops where local appearance cues are already highly discriminative.
In contrast, our continuous conditioning demonstrates more consistent benefits. Among individual attributes, capture year provides the strongest signal, with Year+FiLM achieving the best overall performance (65.9 mAP on head crops and 62.6 on body crops). This indicates that temporal progression is a key factor in longitudinal fish ReID, and that smooth geometric modulation of representations is more effective than step-wise discretization. While combining all three attributes does not always yield additive gains, the continuous variants generally maintain or improve upon the baseline, highlighting the importance of preserving numerical structure when modeling gradual temporal and morphological changes.

In Table~\ref{tab:timeaware_open_ablation}, the limitations of discretized textual conditioning become even more pronounced. While the discrete variant yields a slight improvement on body crops (42.9 to 44.1 mAP), it reduces performance on head crops (51.4 to 49.2 mAP) and remains consistently inferior to continuous conditioning. In contrast, continuous metadata encoding achieves the strongest results (52.0 mAP on head and 47.0 on body), demonstrating improved robustness under simultaneous identity and temporal distribution shifts. These findings suggest that step-wise discretization of numerical attributes may hinder the modeling of gradual appearance evolution in longitudinal fish ReID, whereas preserving the continuous structure of metadata enables smoother adaptation of the embedding space.

On the other hand, several trends can be observed among the proposed metadata conditioning approaches. 
Temporal metadata emerges as the most consistently useful conditioning signal across datasets and evaluation protocols, indicating that temporal progression is a key factor in longitudinal animal ReID. While the optimal conditioning strategy varies across evaluation settings and visual representations, FiLM-based conditioning can yield substantial improvements when temporal metadata is highly informative (e.g., capture year), whereas sinusoidal encoding often provides more stable gains across different scenarios. Importantly, combining multiple metadata attributes does not always outperform single-attribute conditioning, suggesting that temporal information already captures a substantial portion of the relevant distribution shift. This behavior is not unexpected, as the head-crop and full-body evaluations represent distinct visual representations that likely emphasize different identity cues and may therefore benefit from different conditioning strategies. From a practical perspective, temporal metadata (e.g., capture year or acquisition time) provides a strong default choice when metadata availability is limited. Moreover, the consistent improvements observed across multiple metadata types and conditioning mechanisms suggest that the benefits of metadata conditioning are not tied to a single attribute or encoding strategy. Overall, these results demonstrate that continuous metadata conditioning effectively leverages temporal structure and enhances robustness under distribution shifts. \\

\begin{table}[tb!]
\centering
\caption{Metadata conditioning ablation under time-aware closed-set. 
Bold denotes the best result per crop and metric.}
\label{tab:timeaware_close_ablation}
\resizebox{0.5\textwidth}{!}{%
\small
\begin{tabular}{llcccc}
\toprule
\textbf{Attribute} & \textbf{Approach} & \textbf{Crop} & \textbf{mAP} & \textbf{Rank-1} & \textbf{Rank-5} \\
\midrule
Base & None & Head & 63.0 & 57.9 & 82.9 \\
\rowcolor{lightblue}
     &      & Body & 61.3 & 52.6 & \textbf{81.6} \\
\midrule
All 3  & Discrete~\cite{li2025metawild} & Head & 60.8 & 53.9 & 80.3 \\
\rowcolor{lightblue}
     &      & Body &  60.1 & 54.9 & 79.5 \\
\midrule
Dayseq & Sin & Head & 61.5 & 57.9 & 81.6 \\
\rowcolor{lightblue}
       &     & Body & 58.9 & 53.9 & 78.9 \\
       & FiLM & Head & 62.8 & 59.2 & 77.6 \\
\rowcolor{lightblue}
       &      & Body & 60.3 & 55.3 & 77.6 \\
       & Sin+FiLM & Head & 62.3 & 57.9 & 81.6 \\
\rowcolor{lightblue}
       &          & Body & 59.9 & 52.6 & 76.3 \\
\midrule
Length & Sin & Head & 62.1 & 55.3 & 82.9 \\
\rowcolor{lightblue}
       &     & Body & 60.8 & 54.9 & 79.5 \\
       & FiLM & Head & 63.8 & 60.2 & \textbf{85.5} \\
\rowcolor{lightblue}
       &      & Body & 59.4 & 53.2 & 74.1 \\
       & Sin+FiLM & Head & 61.3 & 53.9 & 84.2 \\
\rowcolor{lightblue}
       &          & Body & 58.7 & 51.6 & 76.0 \\   
\midrule
Year & Sin & Head & 62.2 & 57.9 & \textbf{85.5} \\
\rowcolor{lightblue}
     &     & Body & 61.6 & \textbf{56.6} & 80.3 \\
     & FiLM & Head & \textbf{65.9} & \textbf{60.5} & 84.2 \\
\rowcolor{lightblue}
     &      & Body & \textbf{62.6} & 54.9 & \textbf{81.6} \\
     & Sin+FiLM & Head & 58.1 & 51.3 & 80.3 \\
\rowcolor{lightblue}
     &          & Body & 57.5 & 50.0 & 76.3 \\
\midrule
All 3 & Sin & Head & 61.1 & 56.6 & 82.9 \\
\rowcolor{lightblue}
      &     & Body & \textbf{62.6} & \textbf{56.6} & 80.3 \\
      & FiLM & Head & 62.7 & 55.3 & 80.3 \\
\rowcolor{lightblue}
      &      & Body & 57.1 & 51.3 & 73.7 \\
      & Sin+FiLM & Head & 61.0 & 53.9 & 82.9 \\
\rowcolor{lightblue}
      &          & Body & 58.7 & 51.3 & 76.3 \\
\bottomrule
\end{tabular}}
\end{table}

\begin{table}[tb!]
\centering
\caption{Metadata conditioning ablation under time-aware open-set. 
Bold denotes the best result per crop and metric.}
\label{tab:timeaware_open_ablation}
\resizebox{0.5\textwidth}{!}{%
\small
\begin{tabular}{llcccc}
\toprule
\textbf{Attribute} & \textbf{Fusion} & \textbf{Crop} & \textbf{mAP} & \textbf{Rank-1} & \textbf{Rank-5} \\
\midrule
Base & None & Head & 51.4 & 43.4 & 67.1 \\
\rowcolor{lightblue}
     &      & Body & 42.9 & 35.3 & 55.5 \\
\midrule
All 3  & Discrete~\cite{li2025metawild} & Head & 49.2 & 43.4 & 62.4 \\
\rowcolor{lightblue}
     &      & Body & 44.1 & 37.4 & 58.6 \\
\midrule
Dayseq & Sin & Head & \textbf{52.0} & 45.7 & 67.1 \\
\rowcolor{lightblue}
       &     & Body & 45.4 & 39.9 & 59.5 \\
       & FiLM & Head & 47.8 & 42.2 & 60.1 \\
\rowcolor{lightblue}
       &      & Body & 44.9 & 35.8 & 62.4 \\
       & Sin+FiLM & Head & 50.9 & \textbf{46.2} & 62.4 \\
\rowcolor{lightblue}
       &          & Body & 45.0 & 39.9 & 59.0 \\
\midrule
Length & Sin & Head & 50.4 & 43.9 & 68.8 \\
\rowcolor{lightblue}
       &     & Body & 45.2 & 38.8 & 58.7 \\
       & FiLM & Head & 46.3 & 39.9 & 63.0 \\
\rowcolor{lightblue}
       &      & Body & 44.6 & 36.9 & 57.5 \\
       & Sin+FiLM & Head & 51.4 & 45.1 & 63.6 \\
\rowcolor{lightblue}
       &          & Body & 46.1 & 40.2 & 59.8 \\
\midrule
Year & Sin & Head & 51.3 & 44.5 & \textbf{69.9} \\
\rowcolor{lightblue}
     &     & Body & 43.9 & 37.6 & 59.0 \\
     & FiLM & Head & 48.1 & 41.0 & 65.9 \\
\rowcolor{lightblue}
     &      & Body & 46.8 & 37.6 & \textbf{63.0} \\
     & Sin+FiLM & Head & 50.3 & 44.5 & 66.5 \\
\rowcolor{lightblue}
     &          & Body & 45.3 & 37.0 & 61.3 \\
\midrule
All 3 & Sin & Head & 50.1 & 43.4 & 68.2 \\
\rowcolor{lightblue}
      &     & Body & 46.8 & 40.5 & 61.8 \\
      & FiLM & Head & 47.6 & 43.4 & 61.8 \\
\rowcolor{lightblue}
      &      & Body & 43.6 & 36.4 & 53.8 \\
      & Sin+FiLM & Head & 50.0 & 45.1 & 62.4 \\
\rowcolor{lightblue}
      &          & Body & \textbf{47.0} & \textbf{41.0} & 60.7 \\
\bottomrule
\end{tabular}}
\end{table}

For the following ablation studies, we report results under the closed-set protocol using full-body images. This setting is commonly adopted in animal ReID benchmarks, where identity overlap and full-body observations provide sufficient images per identity for stable comparison of architectural variants. Restricting ablations to this configuration reduces additional variability introduced by identity-disjoint or temporally constrained splits and enables more controlled analysis of model design choices.

\subsubsection{Vision Adapter Ablation}
\label{subsec:abl1}
Table~\ref{tab:adapter_ablation} reports the performance of alternative visual adaptation modules under the closed-set protocol using body images. The evaluated adapters cover a range of representative parameter-efficient adaptation strategies, including both lightweight bottleneck designs and more expressive transformer-based modules. All methods are implemented within the same CLIP backbone and trained under identical optimization settings, ensuring a fair and controlled comparison across adaptation mechanisms.
In detail, the backbone is fixed and identical to that used in the proposed method. Prompt learning is enabled: prompts include learnable context tokens and prompt-order ensembling (End, Middle, Front layouts) as described in Sec.~\ref{subsec_prompt}. Metadata conditioning is not used. This configuration, therefore, isolates the impact of visual adaptation while retaining the text-side prompt optimization of our proposed pipeline.
Among conventional adapter designs, the transformer-based adapter achieves the highest performance (41.20\% mAP), outperforming bottleneck and convolutional variants. 
However, all standard adapters remain substantially below the performance of LoRA-based adaptation. LoRA attains 53.00\% mAP with $r=16$ and $\alpha=16$, significantly surpassing the other modules. Conversely, an excessively large rank ($r=512$, $\alpha=1$) yields severe degradation (33.20\%), suggesting that overly flexible low-rank updates can destabilize pretrained representations.

\begin{table}[tb!]
\centering
\caption{Vision adapter ablation on closed-set body images.}
\label{tab:adapter_ablation}
\begin{tabular}{lc}
\toprule
\textbf{Adapter Variant} & \textbf{mAP (\%)} \\
\midrule
Linear Head                & 38.50 \\
\rowcolor{lightblue}
Bottleneck             & 40.60 \\
1D Convolutional       & 39.50 \\
\rowcolor{lightblue}
2D Convolutional       & 40.00 \\
Transformer   & 41.20 \\
\rowcolor{lightblue}
LoRA ($r=8$, $\alpha=8$)      & 48.30 \\
LoRA ($r=16$, $\alpha=16$)    & \textbf{53.00} \\
\rowcolor{lightblue}
LoRA ($r=512$, $\alpha=1$)    & 33.20 \\
\bottomrule
\end{tabular}
\end{table}

\subsubsection{Prompt Layout Ablation}
\label{subsec:abl2}
Table~\ref{tab:prompt_ablation} evaluates the impact of prompt token placement under the closed-set protocol using body images. 
In this experiment, the backbone is fixed and identical to that used in the proposed method, and visual adaptation is applied either via the Transformer Adapter or LoRA. 
Metadata conditioning is disabled. 
Only the relative placement of the learnable context tokens is varied.
We evaluate the three prompt layouts defined in Sec.~\ref{subsec_prompt}. 
Concretely, the \textit{End}, \textit{Middle}, and \textit{Front} layouts differ in the relative placement of the learnable context tokens with respect to the fixed class and identity tokens. 
While the fixed prefix, class (``fish''), and identity tokens remain unchanged, the context tokens are positioned before, between, or after these tokens according to the respective layout. Across both adaptation strategies, prompt layout influences performance, although differences between individual layouts remain moderate. 
For the Transformer Adapter, the \textit{End} layout achieves the strongest performance (48.2\% mAP), while the \textit{Front} layout performs slightly worse. 
For LoRA, the \textit{Middle} layout yields the best single-layout performance (52.4\% mAP). 
Importantly, ensembling all layouts by averaging normalized embeddings consistently improves stability and achieves the best overall result (53.0\% mAP), confirming that prompt diversity enhances semantic alignment in the shared CLIP embedding space.

\begin{table}[tb!]
\centering
\caption{Prompt layout ablation on closed-set body images. 
Results are reported for both Transformer Adapter and LoRA.}
\label{tab:prompt_ablation}
\begin{tabular}{lcc}
\toprule
\textbf{Prompt Layout} & \textbf{Transformer} & \textbf{LoRA} \\
\midrule
End     & 48.2 & 52.0 \\
Middle  & 48.1 & 52.4 \\
Front   & 47.4 & 52.1 \\
\rowcolor{lightblue}
All layouts (ensemble) & 48.2 & \textbf{53.0} \\
\bottomrule
\end{tabular}
\end{table}

\subsubsection{Ablation of Identity Token}
\label{subsec:abl3}
To evaluate the contribution of the identity token in prompt construction, we remove the fixed identity token ($\mathbf{e}_{\mathrm{id}}^{\,i}$) from the prompt while retaining all other components, including LoRA adaptation and auxiliary identity supervision. Under this ablation, prompts consist only of a prefix, learnable context tokens, and class tokens. 
On the closed-set body protocol, removing the identity token decreases mAP by 0.9\%, Rank-1 by 1.3\%, and Rank-5 by 1.0\%.
Although the magnitude of the drop is moderate, it is consistent across metrics and indicates that identity tokens provide complementary supervision beyond metric learning and auxiliary classification. This supports their role as identity-specific conditioning signals that improve cross-modal alignment rather than acting as proxy-like class representatives.

\subsubsection{Ablation of Auxiliary Identity Supervision}
\label{subsec:abl4}
To assess the necessity of auxiliary identity supervision, we remove the auxiliary classification term $\mathcal{L}_{\mathrm{aux}}$ from the overall objective while retaining dual-space triplet supervision and cross-modal alignment. Under the closed-set protocol with body images, this results in decreases of 6.2\% in mAP, 5.7\% in Rank-1, and 7.9\% in Rank-5. These substantial drops indicate that absolute class-level supervision provides important optimization stability and complementary discrimination beyond metric and cross-modal losses.

\subsubsection{Cross-Modal Loss Ablation}
\label{subsec:abl5}
To evaluate the contribution of the symmetric cross-modal alignment objective, we perform ablations on the image-to-text and text-to-image contrastive terms.
We consider the following variants:
(i) without i2t: $\mathcal{L}_{\mathrm{cm}} = \mathcal{L}_{\mathrm{CCE}}(\mathbf{t}, \mathbf{f}_v^{\mathrm{proj}})$;
(ii) without t2i: $\mathcal{L}_{\mathrm{cm}} = \mathcal{L}_{\mathrm{CCE}}(\mathbf{f}_v^{\mathrm{proj}}, \mathbf{t})$;
(iii) without both: $\mathcal{L}_{\mathrm{cm}} = 0$.
All other training settings remain unchanged.
Removing either direction of cross-modal alignment leads to a performance drop compared to the full model, indicating that both image-to-text and text-to-image supervision contribute to identity discrimination (see Table \ref{tab:cross_modal_ablation}). The largest degradation is observed when both terms are removed, confirming the importance of symmetric cross-modal alignment.

\begin{table}[t]
\centering
\caption{Ablation of cross-modal alignment terms under the closed-set body protocol. See text for variants' explanations.}
\begin{tabular}{lccc}
\toprule
\textbf{Variant} & \textbf{mAP (\%)}  \\
\midrule
Ours w/o i2t          & 51.5  \\
Ours w/o t2i          & 51.9  \\
Ours w/o both         & 49.5  \\
\rowcolor{lightblue}
Ours Full (i2t + t2i) & \textbf{53.0} \\
\bottomrule
\end{tabular}
\label{tab:cross_modal_ablation}
\end{table}

\subsubsection{Ablation of Triplet Supervision}
\label{subsec:abl6}
To examine whether directly optimizing the fused descriptor provides additional benefits, we conduct an ablation comparing different metric supervision strategies under the closed-set body protocol. Specifically, we evaluate four variants:
(i) triplet supervision applied only in the visual space, i.e., $\mathcal{L}_{\mathrm{tri}}^{(v)}(\mathbf{f}_v, y)$;
(ii) triplet supervision applied only in the projected space, i.e., $\mathcal{L}_{\mathrm{tri}}^{(p)}(\mathbf{f}_v^{\mathrm{proj}}, y)$;
(iii) triplet supervision applied to the fused descriptor 
$\mathbf{f}_{\mathrm{concat}} = \mathrm{norm}([\mathbf{f}_v ; \mathbf{f}_v^{\mathrm{proj}}])$, 
where $\mathrm{norm}(\cdot)$ denotes $\ell_2$ normalization after concatenation;
(iv) independent triplet supervision in both spaces, 
$\mathcal{L}_{\mathrm{tri}}^{(v)} + \mathcal{L}_{\mathrm{tri}}^{(p)}$ (our default setting).

\begin{table}[t]
\centering
\caption{Ablation on triplet supervision on closed set body images.}
\label{tab_triplet}
\begin{tabular}{l c}
\toprule
\textbf{Triplet Supervision Variant} & \textbf{mAP (\%)}\\
\midrule
Triplet($\mathbf{f}_v$) &  47.6\\
Triplet($\mathbf{f}_v^{\mathrm{proj}}$) &  48.4\\
Triplet($\mathbf{f}_{\mathrm{concat}}$) &  50.7\\
\rowcolor{lightblue}
Triplet($\mathbf{f}_v$ + $\mathbf{f}_v^{\mathrm{proj}}$, Ours) & \textbf{53.0} \\
\bottomrule
\end{tabular}
\end{table}

Table~\ref{tab_triplet} shows that independent supervision of both embedding spaces yields the best performance. Applying triplet loss only in either $\mathbf{f}_v$ or $\mathbf{f}_v^{\mathrm{proj}}$ leads to lower accuracy, while directly supervising the fused descriptor improves performance but remains inferior to dual-space supervision. The proposed strategy achieves a gain of 2.3\% mAP over fused-descriptor supervision, indicating that enforcing metric constraints separately in the raw and projected spaces produces a more discriminative fused representation at inference.

\subsubsection{Temporal Distance and Identification Performance}
\label{subsec:temporal}
To quantify how identification probability varies with temporal separation under the proposed method, we modeled the probability of successful retrieval as a continuous function of temporal distance between observations. Analyses were restricted to different-day comparisons to exclude trivial same-capture matches. Temporal distance was defined as the absolute difference in days between the query image and its corresponding true match. Because the relationship between temporal separation and identification probability was nonlinear, temporal distance was log-transformed using $\log(1 + \text{days})$ before model fitting.

For each evaluation protocol (CS, OS, TACS, and TAOS), we fitted separate binomial logistic regression models describing identification probability as a function of temporal separation. Success was defined independently for Rank-1 and Rank-5 retrieval. The predicted probabilities and corresponding 95\% confidence intervals were obtained from the fitted models and visualized across the empirically observed temporal range of each protocol. Because TACS and TAOS involve shorter temporal intervals by design, prediction ranges were restricted to the observed time span within each configuration.

\begin{figure*}[tb!]
    \centering
    \includegraphics[width=1\linewidth]{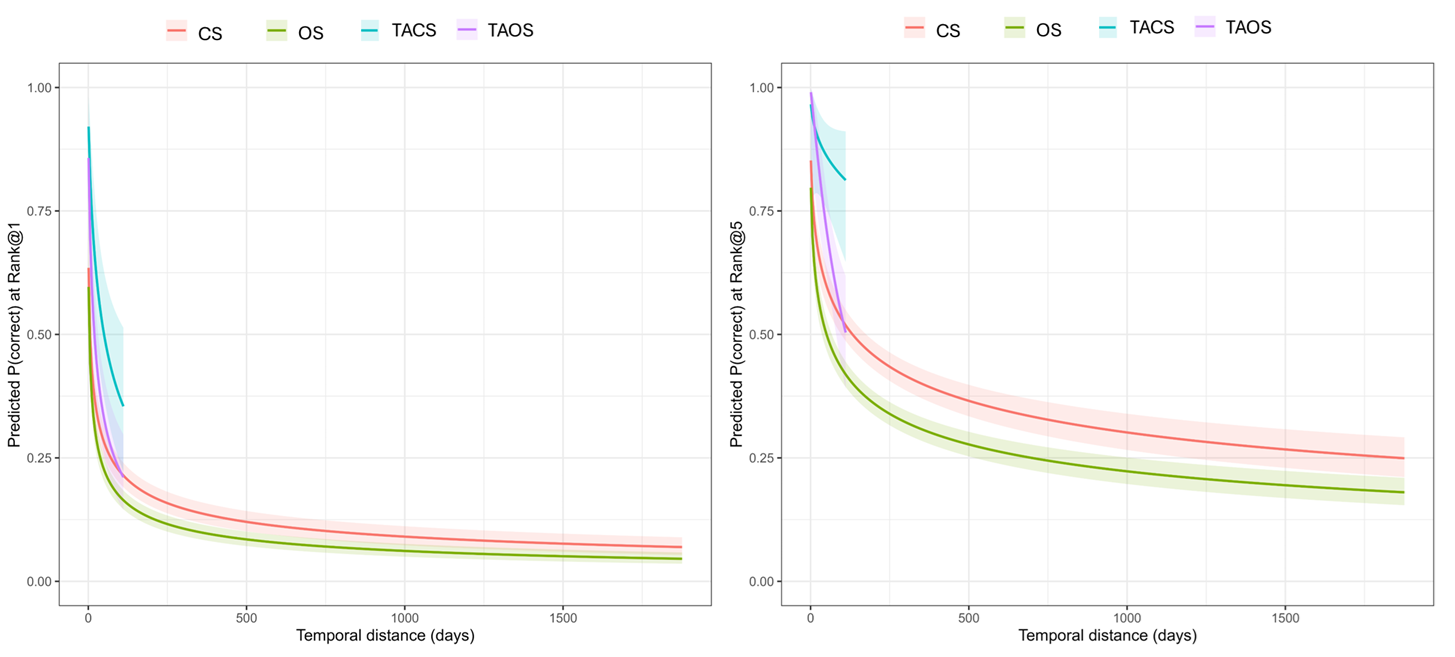}
    \caption{Predicted identification probability as a function of temporal distance under the proposed method. Curves represent fitted binomial logistic regression models for Rank-1 and Rank-5 retrieval across evaluation protocols (CS, OS, TACS, and TAOS). Shaded regions denote 95\% confidence intervals. Temporal distance is log-transformed as $\log(1 + \text{days})$. Prediction ranges are restricted to the empirically observed temporal span of each protocol.}
    \label{fig:temporalAnalysis}
\end{figure*}

ReID accuracy declined significantly with increasing temporal separation across nearly all model configurations (logistic regression with log-transformed temporal distance; all $p < 0.001$ for Rank-1 and Rank-5, except Rank-5 in the TACS protocol, where the effect was not significant). The estimated slopes were negative in all protocols (Rank-1: $\beta_1$ ranging from $-0.46$ to $-0.77$; Rank-5: $\beta_1$ ranging from $-0.42$ to $-1.15$), confirming decreasing identification probability with increasing time gap. Identification probability changed most rapidly at shorter temporal intervals and varied more gradually over longer time scales (Figure~\ref{fig:temporalAnalysis}). Models incorporating temporal constraints achieved higher short-term accuracy compared with the standard CS and OS approaches, although these models were evaluated over shorter temporal ranges.

These results provide direct empirical evidence that ReID performance varies systematically with real temporal separation, indicating that longitudinal variation is measurable and exhibits systematic temporal structure in retrieval performance. The nonlinear decay pattern further suggests that performance changes are more pronounced at shorter temporal intervals, while longer-term variation manifests more gradually. This observation reinforces the relevance of temporally structured evaluation protocols, as the CS/OS and TACS/TAOS configurations exhibit distinct baseline levels and temporal ranges. 

Moreover, the continuous nature of the observed decay supports the conceptual motivation for modeling temporal attributes without discretization, in contrast to discretization-based approaches such as~\cite{li2025metawild}. Because performance varies smoothly as a function of time, conditioning strategies that preserve numerical structure, as applied by the proposed method, are conceptually better suited to capture the underlying longitudinal dynamics than step-wise temporal binning. While this analysis does not directly compare methods, it substantiates the importance of explicitly accounting for temporal variation in long-term animal ReID.

\begin{figure*}[tb!]
    \centering
    \includegraphics[width=0.8\linewidth]{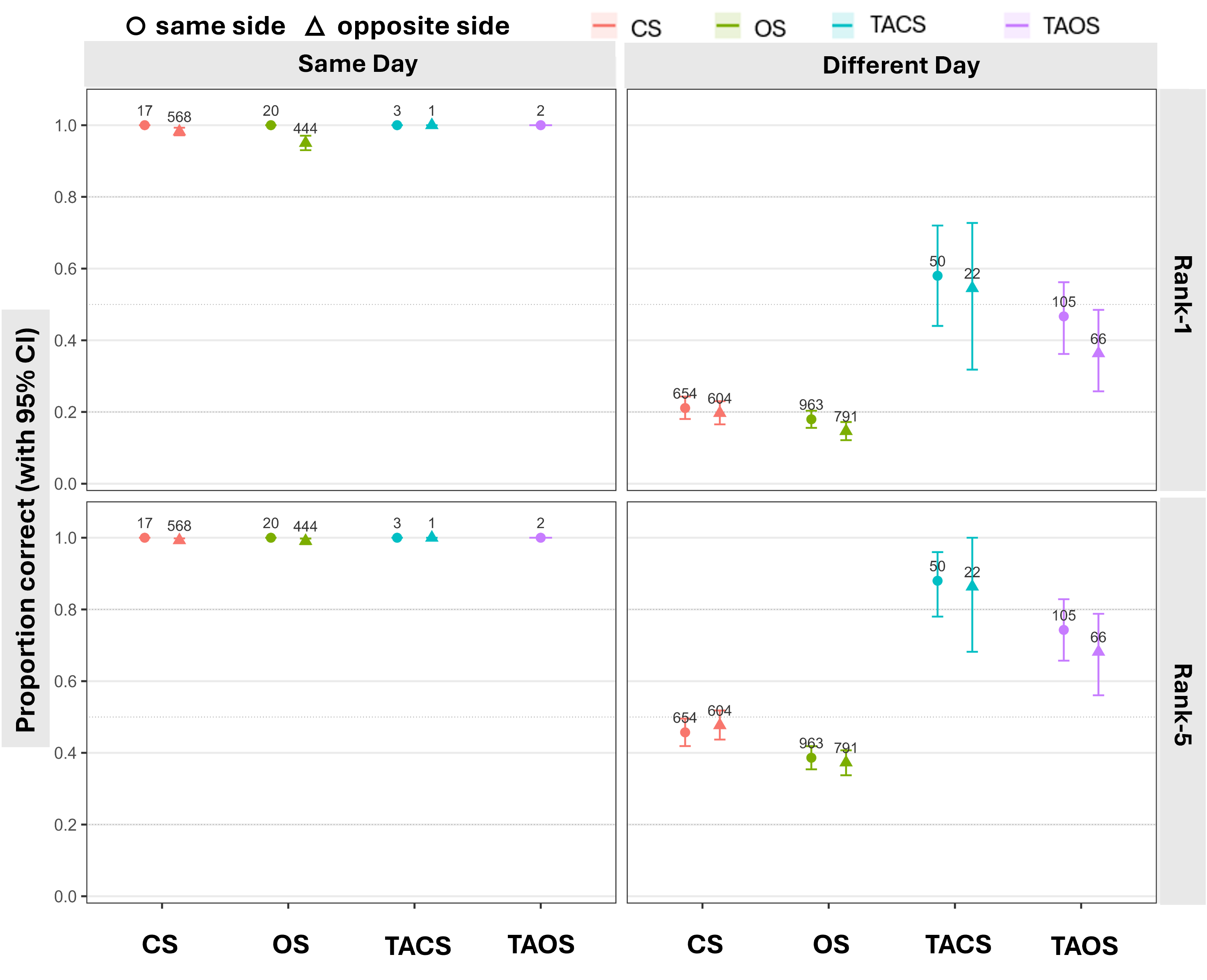}
    \caption{ReID performance across model configurations, stratified by temporal context and body side. Points represent the proportion of queries for which the correct individual was retrieved within the top candidate identities (Rank-1 and Rank-5), with whiskers indicating 95\% bootstrap confidence intervals. Results are shown separately for same-day and different-day comparisons and for matches involving the same body side or opposite body sides. Numbers above points indicate the number of queries contributing to each estimate.}
    \label{fig:bodySide}
\end{figure*}

\subsubsection{Analysis of Temporal and Body-Side Effects}
\label{subsec:stat}

To better understand the impact of temporal variation and body-side differences on ReID performance, we analyze performance across stratified conditions (Fig. \ref{fig:bodySide}). In particular, we consider matches grouped by temporal proximity (same-day vs. different-day) and by body side (same-side vs. opposite-side), motivated by the pronounced left–right asymmetry observed in fish appearance.

ReID accuracy is consistently high for same-day comparisons across all models (Rank-1 $\ge$ 0.95 and Rank-5 $\ge$ 0.99 in most cases), indicating that identity-specific visual cues remain stable over short temporal intervals. However, accuracy declines substantially for different-day matches, reflecting the challenge of longitudinal appearance variation.

Under these conditions, models incorporating temporal constraints achieve noticeably higher identification accuracy than standard closed-set and open-set approaches. For example, Rank-1 accuracy increases from approximately 0.21 (CS) and 0.18 (OS) to 0.58 (TACS) and 0.47 (TAOC) for same-side comparisons, with similar improvements observed for Rank-5. This demonstrates that incorporating temporal structure improves robustness to appearance changes over time.

We note that time-aware models are evaluated over shorter temporal intervals due to constrained candidate sets (1–113 days), whereas CS and OS models operate across the full temporal range of the dataset (1–1877 days). This difference highlights the importance of considering temporal distribution when interpreting performance across protocols.

Finally, differences between same-side and opposite-side matches are relatively small and largely overlap within confidence intervals. This suggests that the model learns features that are robust to left–right asymmetry, and that cross-side matching does not substantially degrade performance. Overall, these results indicate that temporal variation is the primary factor affecting performance, while body-side differences play a comparatively minor role.

Detailed statistics of query counts and temporal gap distributions for each protocol are provided in the Appendix (Table \ref{table:appendix}).

\subsubsection{Efficiency Analysis}
\label{subsec:pea}
To further evaluate computational efficiency, we compare the number of trainable parameters across full fine-tuning of CLIP-B/16 and CLIP-based ReID methods under the same backbone. We report only the parameters updated during training, excluding frozen components, so that the comparison reflects adaptation overhead rather than total model size.
Table~\ref{tab:parameter} summarizes the trainable parameter counts for full fine-tuning, CLIP-ReID~\cite{li2023clip}, IndivAID~\cite{wu2024individual}, and our method with and without metadata conditioning. Full fine-tuning updates both the vision and text encoders, resulting in approximately 150M trainable parameters. In contrast, CLIP-ReID and IndivAID require updating large portions of the backbone or additional modules.

As shown, our approach requires substantially fewer trainable parameters, as adaptation is restricted to low-rank attention updates, learnable prompt context tokens, and lightweight conditioning layers, while keeping the backbone frozen. This demonstrates that the observed performance gains are achieved through structured parameter-efficient adaptation, providing a favorable trade-off between performance and efficiency compared to full fine-tuning.

While Table~\ref{tab:parameter} reports the number of trainable parameters, it also reflects the computational efficiency of the proposed approach. By restricting adaptation to low-rank updates and prompt parameters, the method reduces training memory usage and optimization cost compared to approaches that fine-tune the full backbone.

Importantly, at inference time, all text-related components and metadata conditioning are removed. The model reduces to a standard vision encoder, requiring no additional inputs or modules. Therefore, the inference complexity remains equivalent to the base CLIP model~\cite{radford2021learning}, with no additional computational overhead. For completeness, under our experimental setup (NVIDIA RTX 5090 GPU, batch size = 1), the inference time is approximately 8 ms per image, which is comparable to a standard CLIP ViT-B/16 forward pass.

In addition, compared to multi-stage training strategies such as CLIP-ReID and IndivAID, the proposed method adopts a unified end-to-end optimization scheme, which avoids sequential training phases and simplifies the overall training procedure.

\begin{table}[tb!]
\centering
\caption{Trainable parameter comparison under the same CLIP-B/16 backbone. For metadata-conditioned variants, we report the largest configuration (all metadata attributes), providing an upper bound on parameter count; configurations using fewer attributes are correspondingly more lightweight.}
\label{tab:parameter}
\begin{tabular}{l c}
\toprule
Method & Trainable Params (M) \\
\midrule
Full FT (CLIP-B/16)~\cite{radford2021learning} & 150.00 \\
CLIP-ReID~\cite{li2023clip} & 154.30 \\
IndivAID~\cite{wu2024individual} & 154.40 \\
\rowcolor{lightblue}
Ours & 72.90 \\
\rowcolor{lightblue}
Ours + Meta (All 3 +Sin) & 72.95 \\ 
\rowcolor{lightblue}
Ours + Meta (All 3 +FiLM) & 98.20 \\ 
\bottomrule
\end{tabular}
\end{table}

\subsubsection{Experiments on Other Datasets}
\label{subsec:others}
To further evaluate the generalization ability of our approach across diverse species and acquisition conditions, we conduct additional experiments on multiple publicly available animal ReID datasets. All seven datasets are accessed and standardized using the WildlifeDatasets toolkit~\cite{cermak2024wildlifedatasets,cermak2024wildfusion}, which provides unified data loading, preprocessing, and evaluation protocols (i.e., CS).

To specifically assess cross-dataset generalization in the presence of temporal metadata, we further evaluate our method on the SeaTurtleID2022 dataset~\cite{adam2024seaturtleid}, which spans 13 years and includes timestamp annotations. This dataset differs from the Melops dataset in terms of species, imaging conditions, and acquisition setup, while sharing the availability of temporal metadata and time-aware evaluation protocols (i.e., TACS and TAOS). 
We adopt the same experimental setting and evaluate our method using metadata (year) conditioning with Sin and FiLM configurations.
We use full-body images with background removed using SAM2 \cite{ravi2024sam}. Specifically, we apply segmentation to isolate the turtle and replace the background with a uniform (white) mask, followed by resizing to match the CLIP input resolution. This preprocessing reduces background bias and focuses the model on identity-relevant visual features. We evaluate performance using our standard retrieval protocol (mAP) to ensure consistency with the rest of the paper.

All hyperparameters are kept identical to those used for the Melops dataset, without any dataset-specific tuning. The corresponding mAP results are summarized in Table~\ref{tab:wildlife_generalization}.

For clarity, the primary comparison scope of this work is CLIP-based and parameter-efficient adaptation methods operating under comparable backbone capacity and training regimes.
We do not perform a direct numerical comparison with MegaDescriptor~\cite{cermak2024wildlifedatasets}, as it was trained jointly on 29 wildlife ReID datasets and fully fine-tuned under a large-scale multi-dataset setting. WildFusion~\cite{cermak2024wildfusion} similarly relies on MegaDescriptor as its global backbone, which was trained on over 30 wildlife datasets. In contrast, our approach adapts a frozen CLIP-B backbone using dataset-specific supervision, with training performed individually per dataset under the same-species evaluation protocol. Moreover, MegaDescriptor-L and DINOv2 employed in these studies, use substantially larger transformer architectures than the CLIP-B backbone used in our experiments. Therefore, to ensure a fair comparison under comparable backbone capacity and training regimes, we restrict our evaluation to CLIP-based ReID baselines. Note that when implemented without environmental metadata, the method proposed in~\cite{li2025metawild} reduces to the CLIP-ReID baseline~\cite{li2023clip}, as the metadata-conditioned components become inactive.

As shown in Table~\ref{tab:wildlife_generalization}, our method achieves the best mAP on six out of seven benchmarks and outperforms both CLIP-ReID and IndivAID on the majority of datasets for the CS setting. In particular, noticeable improvements are observed on SeaStarReID2023 (+2.68\% over CLIP-ReID), PolarBearVidID (+1.50\%), and SealID (+2.41\%). These results demonstrate that our parameter-efficient adaptation and prompt-learning strategy maintains strong cross-dataset generalization performance despite operating under a per-dataset training regime without large-scale aggregated supervision.

On the other hand, results on the TACS and TAOS protocols on the SeaTurtleID2022 dataset further validate the effectiveness of metadata conditioning under temporal distribution shifts. Under the TACS setting, incorporating temporal metadata improves performance from 39.26 mAP to 41.13 mAP with sinusoidal conditioning and to 43.37 mAP with FiLM, corresponding to gains of +1.87 and +4.11 mAP, respectively. Similarly, under the more challenging TAOS setting, performance increases from 31.05 mAP to 32.63 mAP (+1.58) with sinusoidal conditioning and to 34.51 mAP (+3.46) with FiLM. These improvements are consistent across both evaluation protocols, demonstrating that metadata conditioning enhances robustness to temporal variation and unseen identities. Notably, these gains are achieved using the same training protocol and hyperparameters as in the Melops dataset, without dataset-specific tuning, highlighting the generalization capability of the proposed approach.

\begin{table}[tb!]
\centering
\caption{Comparisons with CLIP-based methods across various datasets. “N/A” indicates that metadata conditioning is not applied to the corresponding methods, as it is not part of their original design.}
\label{tab:wildlife_generalization}
\resizebox{0.99\linewidth}{!}{
\begin{tabular}{lccc}
\toprule
\textbf{Dataset} & CLIP-ReID & IndivAID & \textbf{Ours} \\
& \cite{li2023clip} & \cite{wu2024individual} &  \\
\midrule
\rowcolor{lightblue}
CS - WildlifeDatasets \cite{cermak2024wildlifedatasets,cermak2024wildfusion}           
& 
& 
&  \\
FriesianCattle2015v2 
& 58.18 
& 61.27 
& \textbf{62.47} \\

AerialCattle2017  
& \textbf{60.09} 
& 57.27 
& 58.18 \\
ATRW              
& 58.05 
& 58.18 
& \textbf{59.52} \\

SeaStarReID2023   
& 57.41 
& 58.18 
& \textbf{60.09} \\
LionData          
& 59.32 
& 58.18 
& \textbf{59.45} \\

PolarBearVidID    
& 58.68 
& 56.70 
& \textbf{60.18} \\
SealID            
& 56.77 
& 58.56 
& \textbf{59.18} \\

\rowcolor{lightblue}
TACS - SeaTurtleID2022 \cite{adam2024seaturtleid}  
& 
& 
&  \\
wout/ Meta            
& 32.44
& 35.73
& \textbf{39.26} \\
w/ Meta (Year+Sin)   
& N/A
& N/A
& \textbf{41.13} \\
w/ Meta (Year+FiLM)   
& N/A
& N/A
& \textbf{43.37} \\

\rowcolor{lightblue}
TAOS - SeaTurtleID2022 \cite{adam2024seaturtleid}        
& 
& 
&  \\

wout/ Meta            
& 29.46
& 30.06
& \textbf{31.05} \\
w/ Meta (Year+Sin)   
& N/A
& N/A
& \textbf{32.63} \\
w/ Meta (Year+FiLM)   
& N/A
& N/A
& \textbf{34.51} \\
\bottomrule
\end{tabular}}
\end{table}

\section{Discussions}
\label{sec:discussions}

The experimental results clarify how the proposed framework performs under realistic longitudinal ecological conditions. 
The absolute performance levels reflect the intrinsic difficulty of longitudinal ecological ReID in the Melops dataset, which comprises nearly 10,000 unique individuals with a highly imbalanced, long-tailed identity distribution. The majority of individuals are observed only once, while relatively few are repeatedly recaptured across years. This sparse per-identity sampling, combined with extended temporal drift and fine-grained inter-individual variation, makes identity discrimination substantially more challenging than in conventional ReID benchmarks. Importantly, such imbalance and limited observations per individual are not artifacts of dataset design, but inherent characteristics of real-world capture–mark–recapture ecological monitoring.

Against this backdrop of extreme sparsity and temporal drift, the ablation analyses clarify how the individual components of the proposed framework contribute to retrieval performance. In particular, they show how low-rank attention adaptation, symmetric cross-modal supervision, and temporal variability jointly influence performance across extended time gaps. The vision adapter ablation demonstrates that moderate-rank low-rank updates to attention projections are sufficient to specialize pretrained CLIP representations for fine-grained identity discrimination without full backbone fine-tuning. Identity adaptation therefore, benefits from limited-capacity updates that preserve pretrained embedding structure, whereas excessively flexible updates reduce discriminability.

Prompt-based conditioning complements low-rank visual adaptation. Although individual prompt layouts yield similar mean performance, prompt-order ensembling improves overall accuracy and reduces sensitivity to token placement, indicating that varying prompt structure enhances cross-modal alignment consistency. The modest but consistent improvement obtained by retaining identity tokens further confirms that identity-conditioned prompts provide supervision beyond auxiliary classification and metric learning. These tokens introduce identity-specific conditioning signals during training while preserving a purely visual inference pipeline at test time. Importantly, unlike proxy-based or prototype-based approaches, identity tokens are not learnable and do not act as class representatives in the embedding space. Instead, they provide non-parametric, identity-specific conditioning within the prompt structure, guiding cross-modal alignment during training without introducing additional representational capacity.

The design of identity-conditioned tokens warrants further clarification. Each training identity is associated with a unique fixed embedding vector that serves solely as a structural conditioning signal within the prompt space. These tokens do not encode semantic attributes and are not optimized during training. Instead, they act as stable anchors that facilitate cross-modal alignment between visual embeddings and identity-specific textual representations. Importantly, identity tokens are used only during training and are discarded at inference time, ensuring that the deployed model remains independent of identity-specific parameters. Keeping identity tokens fixed, rather than learnable, prevents the text branch from absorbing discriminative capacity that should be attributed to the visual encoder. This encourages adaptation through low-rank visual updates and prompt context optimization instead of memorizing identity embeddings in the text space. Scalability is not adversely affected, as identity tokens are instantiated only for identities present in the training split and do not grow with unseen identities at deployment.

The ablation experiments also clarify the relative contribution of training objectives. Auxiliary identity supervision proves essential for stable optimization under parameter-efficient fine-tuning: removing the classification heads produces a substantial performance drop, indicating that absolute class-level supervision complements batch-dependent metric learning and stabilizes optimization of low-rank and prompt parameters. Similarly, symmetric cross-modal alignment provides complementary geometric constraints. Removing either image-to-text or text-to-image contrastive terms reduces accuracy, and removing both leads to the largest degradation, suggesting that bidirectional alignment reinforces structure in the shared embedding space even when the text branch is used only during training. The triplet supervision ablation further shows that applying metric learning independently in both the raw visual space and the CLIP-projected space yields the strongest fused descriptor. This indicates that the two representations capture complementary identity cues and that dual-space supervision is preferable to optimizing a single embedding space or a fused descriptor directly.

Beyond the architectural ablations, the metadata and crop-type analyses offer additional insight into modeling longitudinal variability. Continuous metadata conditioning generally outperforms discretized textual encoding, suggesting that preserving numerical structure is preferable to imposing artificial categorical boundaries when modeling gradual appearance change. Among individual attributes, capture year provides the strongest signal, indicating that coarse temporal progression captures a substantial component of longitudinal drift. The non-additive behavior observed when combining multiple attributes further implies partial redundancy among temporal cues and highlights the importance of selective conditioning rather than indiscriminate feature fusion.

The comparison between head and full-body crops illustrates the role of spatial granularity in identity stability. Full-body cues provide slightly stronger performance under identity-overlapping conditions, whereas head crops exhibit relatively greater robustness in open-set scenarios. This pattern suggests that localized facial features may remain comparatively stable across unseen individuals, while global morphology contributes discriminative information when identity overlap exists but may be more sensitive to growth-related variation. Together, these findings emphasize that both temporal and spatial factors shape representation stability and should be jointly considered in longitudinal ecological ReID.

The temporal analysis further underscores that identity similarity in longitudinal animal ReID is inherently time-dependent. The consistent decline in retrieval probability with increasing temporal separation indicates that temporal drift constitutes a systematic source of distribution shift rather than random noise. Because each evaluation protocol is constructed from distinct splits with different identity compositions and temporal ranges, absolute performance levels are not directly comparable across configurations.
In addition, the relatively small number of query identities in the time-aware protocols reflects the inherent sparsity of long-term recaptures in ecological monitoring, and may introduce increased sensitivity to sample variability; however, this setting provides a realistic evaluation of longitudinal robustness and should be interpreted in that context.
Nevertheless, the negative association between elapsed time and identification reliability is observed within all protocols, reinforcing the importance of temporally structured evaluation. The smooth decay pattern suggests that appearance variation evolves continuously rather than in discrete steps, conceptually supporting conditioning strategies that preserve the numerical continuity of temporal attributes. Overall, temporal separation emerges as a fundamental dimension of robustness that should be explicitly considered in longitudinal ecological ReID.

Across additional wildlife benchmarks, the proposed method maintains consistent improvements over CLIP-based baselines while requiring fewer trainable parameters and per-dataset training. This indicates that the observed performance gains arise from structured parameter-efficient adaptation rather than increased model capacity. Although performance varies across datasets due to differences in acquisition conditions and species-specific appearance characteristics, the overall improvements demonstrate that low-rank visual adaptation and prompt-based supervision generalize beyond the fish-specific setting. Taken together, these findings indicate that combining constrained visual adaptation, structured prompt learning, and continuous metadata conditioning provides an effective and computationally efficient framework for longitudinal animal ReID, while highlighting the importance of explicitly modeling time-dependent appearance variation in ecological monitoring applications.

\section{Conclusions}
\label{sec:conclusion}

In this paper, we presented a parameter-efficient vision–language adaptation framework for longitudinal animal ReID. The framework combines LoRA-based visual adaptation, prompt learning, and cross-modal alignment within a frozen CLIP backbone, while introducing a continuous metadata-conditioning mechanism that incorporates numerical metadata as a training-time-only conditioning signal. Unlike prior approaches that discretize metadata into textual categories or require metadata during deployment, the proposed formulation shapes the embedding geometry during training without introducing any inference-time dependency. Empirical evaluation across multiple protocols and species demonstrates that this conditioning strategy improves robustness under identity and temporal distribution shifts while preserving a purely visual inference pipeline without inference-time metadata or auxiliary components.

Despite these advances, several directions remain for future research. The time-aware protocols involve fewer identities than the full dataset, reflecting the natural sparsity of recaptures in longitudinal ecological monitoring. Such imbalance is inherent to real-world capture–mark–recapture studies, where only a subset of individuals is observed across extended time intervals. While this reduces sample density under strict temporal constraints, it provides a realistic evaluation of long-term identity stability. Future work may therefore focus on extending longitudinal data collection efforts and increasing recapture density, enabling more statistically robust evaluation under strict temporal separation. The present study emphasizes scalar, temporal, and morphological metadata. Extending continuous conditioning to behavioral attributes, such as activity state or reproductive behavior~\cite{canovi2024trajectory}, represents a promising direction. Incorporating such signals would require consistent behavioral annotation in longitudinal datasets but may further improve robustness under complex ecological variation. More broadly, future research may explore continual adaptation strategies for long-term ecological monitoring, cross-species transfer of identity representations, and deeper theoretical analysis of geometric conditioning effects in representation learning.

\section*{Acknowledgment}
This work was supported by the Research Council of Norway (Computer vision to expand monitoring and accelerate assessment of coastal fish (CoastVision), project number 325862.

\newpage

\section*{Appendix}
Table~\ref{table:appendix} provides detailed statistics of query counts and temporal gap distributions across different evaluation protocols. As shown, the closed-set (CS) and open-set (OS) protocols span the full temporal range of the dataset (up to 1877 days), while the time-aware protocols operate over shorter temporal intervals due to constrained candidate sets.

This difference in temporal coverage should be taken into account when interpreting performance comparisons across protocols, as it directly affects the difficulty of the ReID task.

\begin{table}[h]
\centering
\caption{Summary of evaluation conditions across models and temporal groups. The table reports the number of queries ($n$) and the corresponding temporal gap statistics.}
\label{table:appendix}
\begin{tabular}{llcc}
\toprule
Model & Day Group & $n$ & Temporal Gap (days) \\
\midrule
CS & Different day & 1258 & 364 (1--1877) \\
OS & Different day & 1754 & 359 (1--1877) \\
TACS & Different day & 72 & 54.8 (1--113) \\
TAOS & Different day & 171 & 56.3 (1--113) \\
\midrule
CS & Same day & 585 & 0 \\
OS & Same day & 464 & 0 \\
TACS & Same day & 4 & 0 \\
TAOS & Same day & 2 & 0 \\
\bottomrule
\end{tabular}
\end{table}

\bibliographystyle{cas-model2-names}

\bibliography{references}



\end{document}